\documentclass[10pt,journal]{IEEEtran}
\usepackage{graphicx,color,multicol,url,amssymb,multirow,amsmath,caption}
\usepackage[table]{xcolor}
\newcommand\footnoteref[1]{\protected@xdef\@thefnmark{\ref{#1}}\@footnotemark}

\usepackage{chngcntr}
\counterwithin*{paragraph}{section}

\makeatother

\newcommand\blfootnote[1]{%
  \begingroup
  \renewcommand\thefootnote{}\footnote{#1}%
  \addtocounter{footnote}{-1}%
  \endgroup
}
\usepackage[width=170mm,left=20mm,paperwidth=210mm,height=239mm,top=20mm,paperheight=279mm]{geometry}
\begin{document}

\title{Are object detection assessment criteria ready for maritime computer vision?}\vspace{-3mm}
%
%

\author{Dilip K. Prasad$^{1,2,*}$, Huixu Dong$^3$, Deepu Rajan$^2$, and Chai Quek$^2$}
\maketitle
\blfootnote{\IEEEcompsocitemizethanks{\IEEEcompsocthanksitem{$^1$Department of Computer Science, UiT The Arctic University of Norway, Troms\o $ $ 9037, Norway, Email-$^*$dilipprasad@gmail.com} }}
\blfootnote{\IEEEcompsocitemizethanks{\IEEEcompsocthanksitem{$^2$School of Computer Science and Engineering, Nanyang Technological University, Singapore 639798} }}
\blfootnote{\IEEEcompsocitemizethanks{\IEEEcompsocthanksitem{$^3$Robotics Institute, Carnegie Mellon University, Pittsburgh, PA 15213} }}

\maketitle

\vspace{-3mm}
\begin{abstract}
Maritime vessels equipped with visible and infrared cameras can complement other conventional sensors for object detection. However, application of computer vision techniques in maritime domain received attention only recently. The maritime environment offers its own unique requirements and challenges. Assessment of the quality of detections is a fundamental need in computer vision. However, the conventional assessment metrics suitable for usual object detection are deficient in the maritime setting. Thus, a large body of related work in computer vision appears inapplicable to the maritime setting at the first sight. We discuss the problem of defining assessment metrics suitable for maritime computer vision. We consider new bottom edge proximity metrics as assessment metrics for maritime computer vision. These metrics indicate that existing computer vision approaches are indeed promising for maritime computer vision and can play a foundational role in the emerging field of maritime computer vision.
\end{abstract}
\vspace{-3mm}

\section{Introduction}

Maritime vessels (MV) are equipped with sensors such as radar, sonar and LIDAR for situational awareness. The automatic identification system (AIS) supports traffic data exchange over maritime communication channels, through which each MV with on-board AIS declares its position, speed, and intended path. The International Regulations for Preventing Collisions at Sea 1972 (COLREGs) impose that all cargo ships weighing more than 300 tonnes and all passenger ships are equipped with AIS. There is no such imposition on smaller MVs, including fishing boats and small-medium sized cargo MVs. Such MVs are invisible in traffic data. Moreover, the AIS channel may be inaccessible for several minutes to few hours at a time \cite{Tu2017}. Cameras in the visible and infrared (IR) range now play a complementary role by overcoming disadvantages of traditional sensors like the minimum range associated with radar and sonar \cite{bloisi2009argos}. Thus, {\textit{computer vision (CV) techniques should play an important role in detecting objects in the maritime environment}}, especially in detecting small and medium sized MVs that have weak radar or sonar signatures and lack on-board AIS.

\begin{figure}[t]
  \centering
  {\small (a) Physical distances vary non-linearly in image \cite{withagen1999automatic,cuzzocrea2017advanced}}\\
    \includegraphics[width=0.9\linewidth]{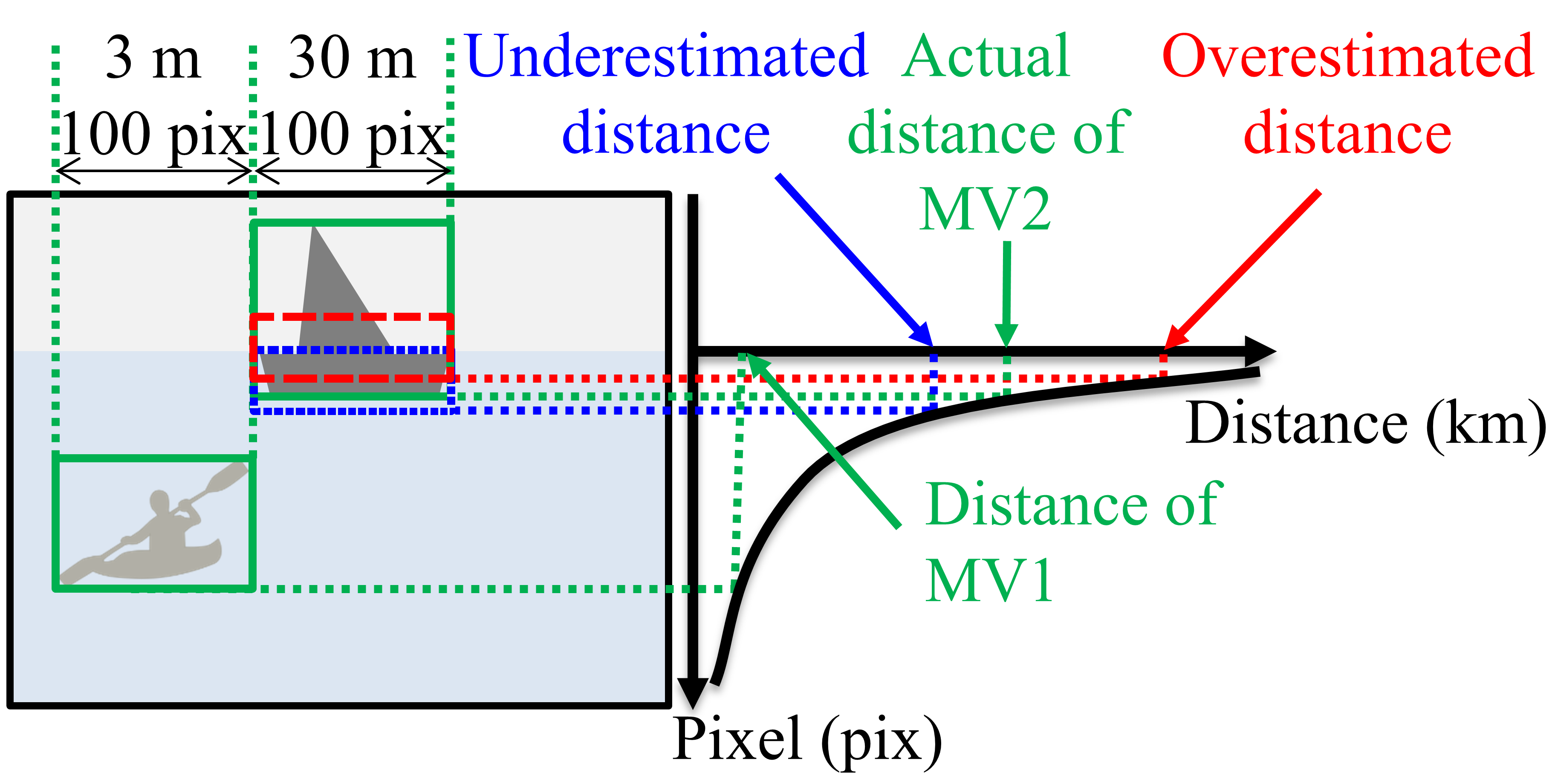}\\
  {\small (b) 10 Examples of maritime objects' appearance}
   \includegraphics[width=1\linewidth]{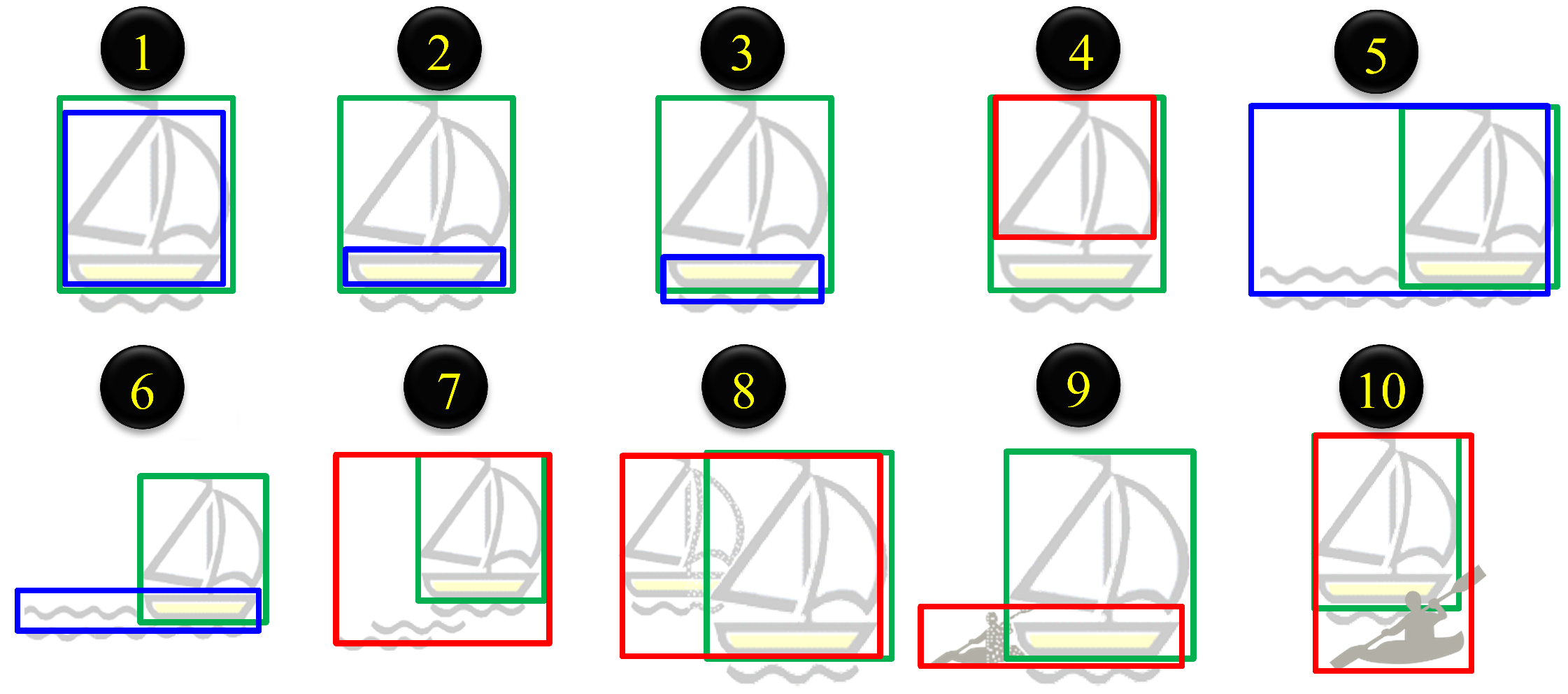}\\\vspace{-3mm}
  \caption{\small What is an acceptable detection of a maritime vessel? (a) Collision avoidance requires accurate estimate of the distance, which is related to the bottom edge of the vessel, and the minimum span of a maritime object. (b) Green, blue, and red boxes denote ground truth, acceptable detection, and unacceptable detections, respectively.}\label{fig:distance} \vspace{-3mm}
\end{figure}

Maritime CV for object detection faces several challenges. Maritime video streams are characterized by {\textit{scene flatness}}, i.e. lack of landmarks and marked lanes as in roads. The maritime scene offers {\textit{difficult to model dynamic background}} features because of challenges such as a semi-stochastic wave background, the sharp contrasts of wakes, possibilities of occlusion of MVs, and weather and illumination conditions such as rain, haze and glint \cite{prasad2017TITS}. Further, planning the manoeuver and deceleration for collision avoidance (CA) is challenging since the distance and span of the MVs in the scene is related non-linearly to the pixels along the $y-$axis \cite{withagen1999automatic,cuzzocrea2017advanced}, see Fig. \ref{fig:distance}(a). {There are other applications also, that face the same non-linearity between the physical space and image space. For example, a reviewer of this manuscript suggested terrestrial applications, where ``obstacle detection by automative vehicle sensors (for automated braking for example) has the same bias, since the flat world assumption is usually used in this domain too.''} An appropriate maritime CV solution has to satisfy the following requirements:
\begin{itemize}
  \item detect and track MVs in the scene
  \item determine MVs' accurate spans, positions and tracks
  \item provide real-time results
  \item perform in all weather and illumination
\end{itemize}

Detection and tracking of MVs falls under the ensemble problem set of `detection and tracking in a dynamic background', which has been extensively studied in computer vision. The existing CV solutions in this ensemble can provide a firm foundation for developing dedicated CV solutions for maritime object detection requirements. {We note that the above identified goals of maritime CV comprise a broad topic and entail research for several years to come. In this paper, we choose a very specific problem within this broad scope and critical for the entailing research. The specific problem considered in this is paper as follows.} Adoption of existing CV solutions for maritime CV encounters a set back. We show that traditional performance measures for object detection fail in the maritime environment and we discuss the following question. {\textit{How do we assess the quality of detection for maritime computer vision?}}


We show that assessment metrics such as intersection over union (IOU, also called Jaccard index \cite{Levandowsky1971}) and intersection over ground truth (IOG, also called sensitivity \cite{Altman1994}), most often used in object detection, are unsuitable for maritime CV. They are deficient in assessing the accuracy of span and distance of detected MVs. Either the detection method provides a very high IOU, say 90\%, or customized assessment metric is needed to meet the requirements of maritime CV. {\textit{The aim of this paper is to design custom assessment metrics that provide good assessment of the quality of detected objects while not putting severe demands on detection algorithms.}}

We discuss two new assessment metrics customized for maritime computer vision. We also study the performance of existing background subtraction (BGS) algorithms and regions with convolution neural network (R-CNN) features using conventional and proposed assessment metrics. We show that the conventional metrics indicate general unsuitability of BGS algorithms for maritime CV whereas the new metrics present hope of using them in maritime CV. We expect that this exercise shall provide useful cursors for developing maritime CV solutions.

The assessment requirements of maritime CV are discussed in section \ref{sec:requirement}. The deficiency of conventional metrics for maritime CV is discussed in section \ref{sec:conventional}. The proposed bottom edge proximity metrics are presented and compared with conventional metrics in section \ref{sec:proposed}. Experimental results of existing BGS algorithms and R-CNN on a maritime dataset are presented in section \ref{sec:results}. Section \ref{sec:discussion} concludes this paper with a discussion on the future outlook for maritime CV.

\begin{figure}[t]
  \centering
  \includegraphics[width=\linewidth]{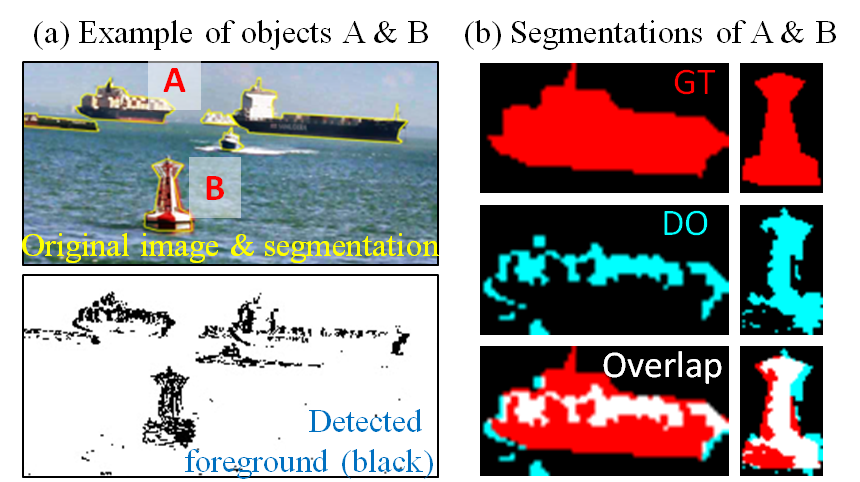}\\
  \small (c) Other example objects C-I from 5 different videos\\
  \includegraphics[width=0.95\linewidth]{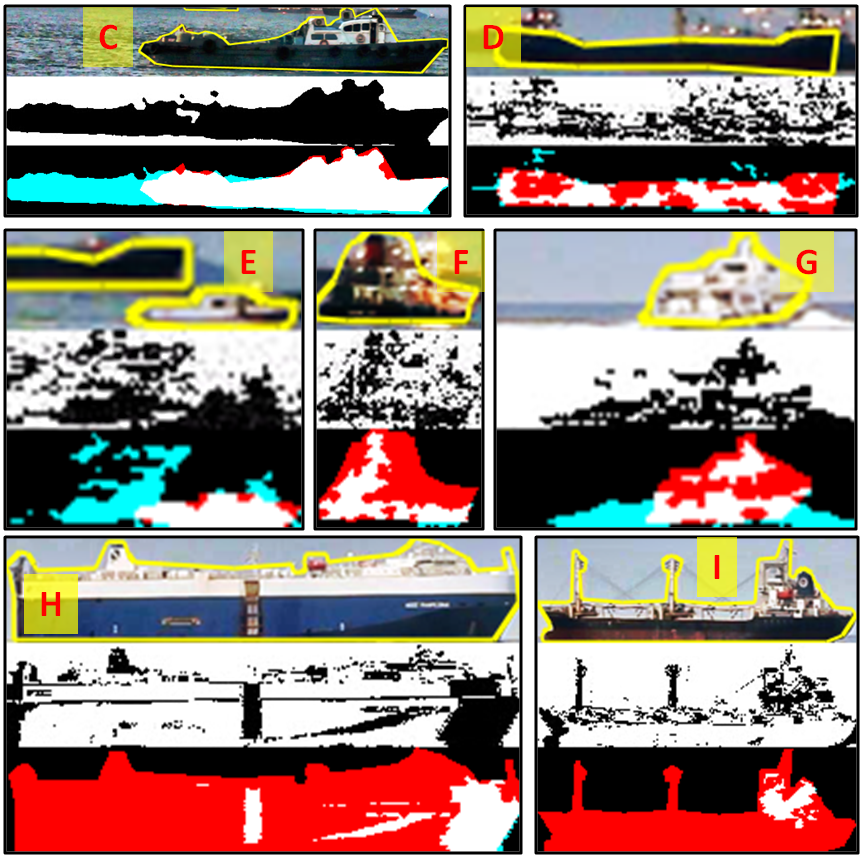}\\
  \small (d) Comparison of pixel-based and BB based segmentations\\ \vspace{1mm}
  \scalebox{0.9}
  {
  \begin{tabular}{|c|c|c|c|c|}
    \hline
    Object & \multicolumn{2}{c|}{IOU} & \multicolumn{2}{c|}{IOG}\\
    \cline{2-5}
    & Pixel-based & BB-based & Pixel-based & BB-based \\ \hline
     A  & 0.27 & 0.64
              & 0.29 & 0.83 \\
     B  & 0.55 & 0.85
              & 0.64 & 0.85 \\
     C  & 0.68 & 0.65
              & 0.95 & 0.96 \\
     D  & 0.45 & 0.60
              & 0.49 & 0.97 \\
     E  & \textbf{0.39} & \textbf{0.22}
              & 0.87 & 1.00 \\
     F  & 0.39 & 0.89
              & 0.40 & 0.99 \\
     G  & 0.36 & 0.42
              & 0.47 & 0.91 \\
     H  & 0.15 & 0.47
              & 0.15 & 0.48 \\
     I  & 0.09 & 0.10
              & 0.09 & 0.10 \\
    \hline
  \end{tabular}
  }
  \caption{\small Pixel segmentations are more demanding than bounding boxes, see subfigures (a-c) for qualitative examples. The same metrics result into significantly lower values when computed for pixel segmentations, as noted in subfigure (d). The only exception in the examples considered in (a-c) is shown in bold in (d). In the subfigure (c), there are 3 panels for each object. The top, middle, and bottom panels respectively show GT in image, detected foreground pixels, and the overlap of DO and GT. }\label{fig:pixel}
  \vspace{-4mm}
\end{figure}

\section{Requirements for maritime CV}\label{sec:requirement}

Before discussing the suitability of conventional metrics, or lack thereof, we consider the fundamental question: `What is an acceptable detection of a maritime vessel?'. It is important to accurately estimate the location of the MV in a scene (given by the bottom edges of the MV) and its minimum span (determined by the width of the MV in pixels and its position in the image frame). See Fig. \ref{fig:distance}(a) for illustration. Consider the example cases 1-10 shown in Fig. \ref{fig:distance}(b). Example 1 is close to ideal, where the bounding box (BB) of the detected object (DO) is almost the same as the BB of the ground truth (GT). We restrict our discussion to bounding boxes because the pixel-segmentations are significantly more demanding than bounding boxes. We illustrate this point using Fig. \ref{fig:pixel}. Fig. \ref{fig:pixel}(a) shows two objects A and B in an image and also the foreground segmentation result obtained using a dynamic background subtraction method. Their pixel segmentations of GT and DO are shown in Fig. \ref{fig:pixel}(b). Other examples are shown in Fig. \ref{fig:pixel}(c). Fig. \ref{fig:pixel}(d) shows values of IOU and IOG for pixel and BB segmentations. The small values of IOU and IOG for the pixel segmentations of almost all the objects indicate that assessing the pixel segmentations is more demanding. Moreover, the pixel segmentations are not particularly more informative than BBs about the distance and span of the vessel anyway. Yet, at least for one example, i.e. object E in Fig. \ref{fig:pixel}(c,d), the IOU for pixel based segmentation is larger than BB segmentation. Therefore, the importance of pixel segmentations in accurate detection of MVs cannot be discounted. It merits an elaborate study, which we relegate to the future work.

Although there is a large variety of MVs, in general, an MV is characterized by a hull and an optional super-structure, i.e. all parts above the hull, including masts. The existing CV solutions may detect hull and super-structure separately due to two reasons. First, super-structure is not an essential component and supervised learning approaches may undertrain for vehicles with super-structures. Second, stark differences in geometries, color, and other image features of the hull and the super-structure imply that the super-structure may appear as an independent object. The hull or the super-structure may even be left undetected, such as in the case of sailboats, due to a lack of contrast between the background and the super-structure. Consequently, the DO may appear as shown in examples 2-4. {\textit{For collision avoidance, accurate detection of the hull is important, irrespective of whether the super-structure is included in the DO with the hull (example 1), detected independently (example 4), or not detected at all (examples 2 and 3).}} Furthermore, the physical distance between the MV and the sensor is mapped non-linearly in an image along a direction perpendicular to the horizon (see Fig. \ref{fig:distance}(a)). This means that the line in image corresponding to horizon is at infinity while the bottom most pixel is only a few meters away from the sensor. Thus, {\textit{incorrect estimation of bottom of hull may result in hugely incorrect estimation of the physical distance}}. However, it is preferable to slightly underestimate the distance between the sensor and an MV for collision avoidance, rather than overestimate it. In this sense, DOs in examples 2 and 3 of Fig. \ref{fig:distance}(b) are acceptable.

Current BGS solutions for object detection struggle with the presence of wakes of maritime vessels \cite{prasad2017TITS}. Often wakes are detected as part of the MVs, such as shown in examples 5-7. Similar to the logic of underestimating the distance between the sensor and the detected MV, it is safer if the estimated width is not lesser than the actual span. Thus, horizontal wakes becoming a part of DO is acceptable, though not preferable. However, {\textit{large extension of the DO in the vertical direction below the hull may result in grossly incorrect estimate of distance, and is not preferred}} (see example 7).

\begin{figure}[t]
  \centering
  \includegraphics[width=1\linewidth]{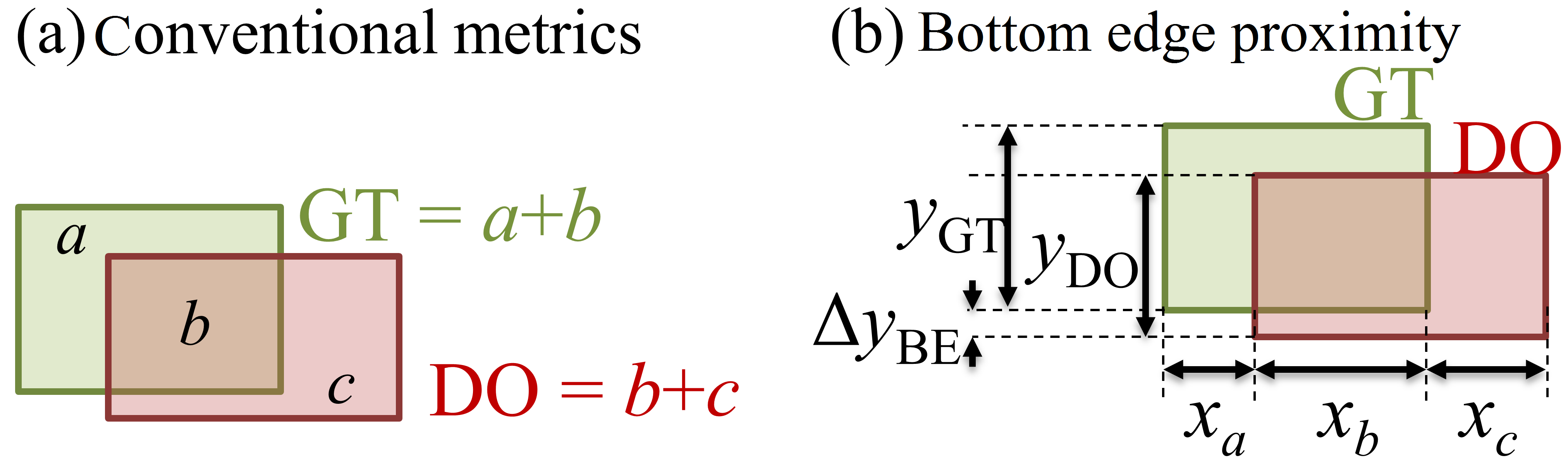}\\ \vspace{-3mm}
  \caption{\small The notations relevant to the conventional metrics and the proposed bottom edge proximity (BEP) metrics are shown here.}
  \label{fig:notations} \vspace{-3mm}
\end{figure}

\begin{figure}[t]
  \centering
  \includegraphics[width=0.7\linewidth]{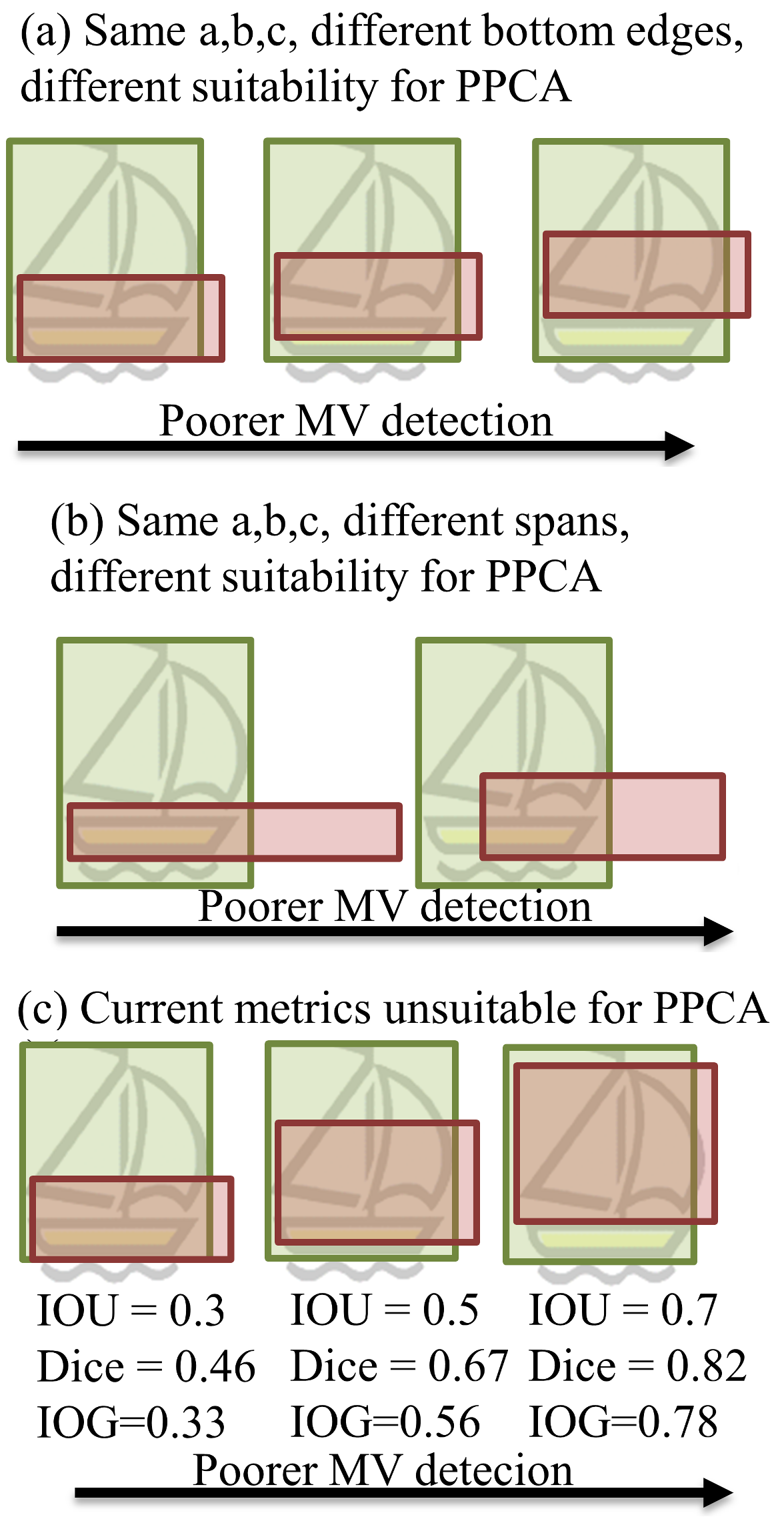}\\ \vspace{-1mm}
  \caption{\small The current metrics are unsuitable for assessing detected objects in maritime CV. For the same values of a, b, and c, one DO may be preferred over others (a,b). Increasing IOU, Dice Index, or IOG metrics need not indicate better detections (c).} \label{fig:currentmetrics} \vspace{-3mm}
\end{figure}

{\textit{The condition of occlusion has a significant implication on collision avoidance. The extension of DO due to occlusion in any direction may mean that the MV with smaller pixel footprint is not detected}} (see examples 8-10). Though the DOs for all these examples are not preferred, the implications are much more severe for examples 9-10, which involve a small MV (kayak) with no on-board communication channel and poor detectability in radar and sonar. These situations call for a close to perfect overlap between the DO and the GT. However, even between examples 9 and 10, example 10 is the least preferred detection. In example 10, the DO leads to gross underestimation of the location of large MV and missed detection of a kayak that is much closer to sensor, much agile, and invisible in other sensor streams.

\begin{figure*}[t]
  \centering
  \includegraphics[width=0.9\linewidth]{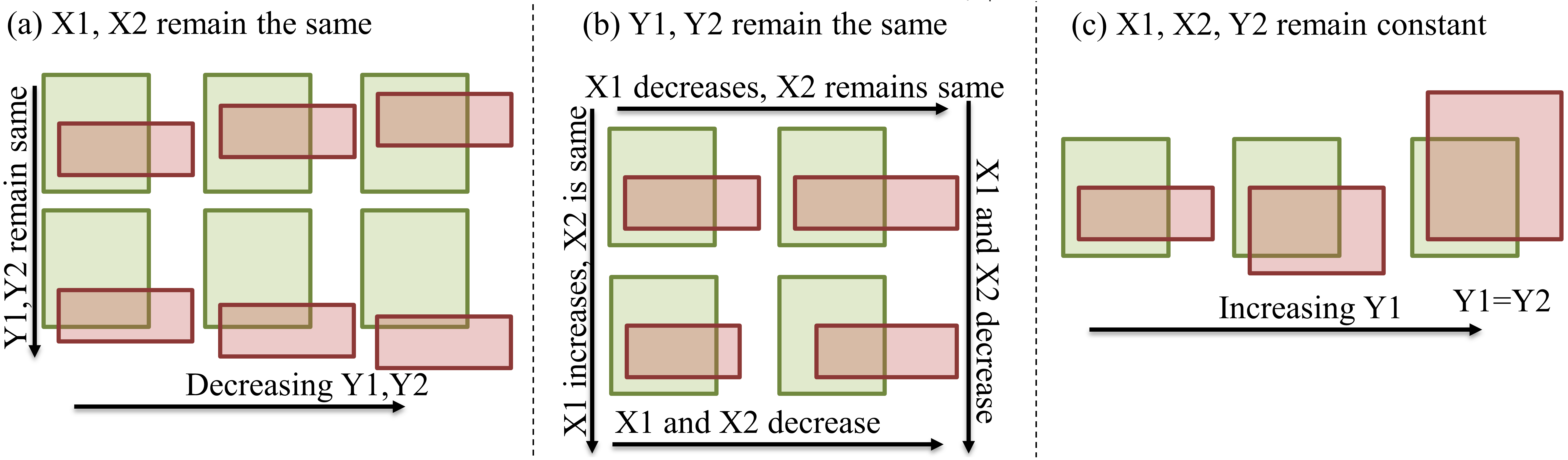}\\ \vspace{-1mm}
  \caption{\small BEP is sensitive to the bottom edges of the DO and GT (a). $X_1$ is more strict than $X_2$ (b). $Y_1$ is more strict than $Y_2$ (c). Thus $\rm BEP_1$ is more strict than $\rm BEP_2$.} \label{fig:BEP} \vspace{-3mm}
\end{figure*}



\section{Conventional assessment criteria versus the needs of maritime CV}\label{sec:conventional}

Assessment of the quality of detection is usually performed through similarity metrics, such as Jaccard index \cite{Levandowsky1971} (also called IOU) or Dice index \cite{Dice1945}. Their generalized form is given by Twersky index \cite{tversky1977features}, defined as follows:

\begin{equation}\label{eq:IOU}
{ \rm S
= \small{\frac{b}
  {b+\alpha a+\beta c}
  }}
\end{equation}
where $a, b, c$ are the areas of $\rm (GT - DO)$, $\rm (GT \cap DO)$, and $\rm (DO - GT)$, respectively (see Fig. \ref{fig:notations}(a)). The parameter $\alpha$ emphasizes the allegiance of the overlapped region with GT while the parameter $\beta$ emphasizes the allegiance of the overlapped region with DO. Similarity metrics usually employ symmetry with respect to GT and DO, i.e. $\alpha=\beta$. Dice index corresponds to $\alpha = \beta = 0.5$ and widely used IOU corresponds to $\alpha = \beta = 1.0$. A detection is assessed as true positive if IOU $>c_0$. Similar threshold is employed if other similarity metrics are used. Usually in CV, IOU $>$ 0.5 is considered sufficient. We consider an additional asymmetric metric with $\alpha=1$, $\beta=0$, which we refer to as intersection over ground truth (IOG). This metric assesses the intersection area $b$ with respect to the area of GT ($a+b$) only. Thus excess span detection due to wakes (examples 5-7 in Fig. \ref{fig:distance}(b)) or excess detection in vertical direction below the hull (example 3 in Fig. \ref{fig:distance}(b)) do not affect the assessment negatively if the metric IOG is used.

The essential problem with the above metrics is that two cases may have the same areas $a,b,c$, but one case may be a preferred detection over another. See Fig. \ref{fig:currentmetrics}(a,b) for examples. Also, the increasing value of the above mentioned metrics need not imply better detection, as shown in Fig. \ref{fig:currentmetrics}(c). New metrics that account specifically for the importance of the bottom edge of the hull are needed.

\section{Proposed bottom edge proximity criteria}\label{sec:proposed}

We consider two new criteria that specifically judge the accuracy of detection of the bottom edge (BE) and the span of the DO. We call them bottom edge proximity 1 (BEP$_1$ appears here for the first time) and bottom edge proximity 2 (BEP$_2$, recently proposed in \cite{prasad2018TITS}). BEP$_1$ is symmetric with respect to DO and GT while BEP$_2$ is biased towards allegiance with GT. We use the notations in Fig. \ref{fig:notations}(b) for the definitions of BEP$_1$ and BEP$_2$ presented next.

\paragraph{Bottom edge proximity 1 (BEP$_1$)} We define ${\rm BEP_1} = X_1 Y_1$ where

\begin{equation}\label{eq:X_1}
  X_1 = \frac{x_b}{x_a+x_b+x_c};\,\, Y_1=1-\frac{\Delta y_{\rm BE}}{\min(y_{\rm GT},y_{\rm DO})}
\end{equation}

The smaller the distance between the edges of the GT and DO, the larger is $Y_1$. See Fig. \ref{fig:BEP}(a) for an illustration of this point. However, if the DO is significantly smaller than GT, $Y_1$ becomes poorer. Thus, it indirectly embeds the vertical size of DO in comparison with GT. This is shown in Fig. \ref{fig:BEP}(c).

\paragraph{Bottom edge proximity 2 (BEP$_2$)} We define ${\rm BEP_2} = X_2 Y_2$ where
\begin{equation}\label{eq:X_2}
  X_2 = \frac{x_b}{x_a+x_b};\,\, Y_2=1-\frac{\Delta y_{\rm BE}}{y_{\rm GT}}
\end{equation}

We note that ${\rm BEP_1}$ is stricter than ${\rm BEP_2}$. This is because $X_1$ is less tolerant to extended span of DO due to wakes as well as occlusions, as shown in Fig. \ref{fig:BEP}(b). Further, $Y_1$ is sensitive to the size of DO if the DO is smaller than the GT, as shown in Fig. \ref{fig:BEP}(c).

For convenience, we refer to $X_1$ and $X_2$ as $X$ metrics. Similarly, we refer to $Y_1$ and $Y_2$ as $Y$ metrics. An advantage of BEP metrics is that the threshold(s) for assessing a detection as a true positive can be chosen flexibly. Either a single threshold $c_0$ can be used for the net BEP score, or two thresholds $x_0$ and $y_0$ can be considered for $X$ and $Y$ metrics independently, and a TP can be assessed if both conditions $X>x_0$ and $Y>y_0$ are satisfied.

\paragraph{Qualitative comparison for examples in Fig. \ref{fig:distance}(b)}
We perform a qualitative comparison of the metrics IOU, Dice index, IOG, $\rm BEP_1$, and $\rm BEP_2$ on the examples in Fig. \ref{fig:distance}(b), which were used to study acceptable and unacceptable detections for maritime CV. The results are shown in Table \ref{tab:examples}. We briefly discuss the selection of the thresholds (given in parentheses) for the metrics. Since the threshold value of $c_0=0.5$ is conventionally used in object detection \cite{everingham2015pascal}, we use this value for IOU. Similarly, we use $c_0=0.5$ as threshold for the Dice index and IOG as well. Since $X_1$ and $X_2$ are 1-dimensional analogues of the 2-dimensional IOU and IOG, we use a threshold value of $x_0= \sqrt{0.5}$. Lastly, we use threshold value of $y_0=0.75$ because the accuracy of bottom edge is critical in collision avoidance.

As discussed before, conventional metrics that use $a,b,c$ shown in Fig. \ref{fig:currentmetrics} are not suitable for assessing detections in maritime CV. This is evident in Table \ref{tab:examples}, where IOU, Dice index, and IOG have successes for less than half the number of examples. $\rm BEP_1$ performs better, getting 6 successes out of 10 examples. $\rm BEP_2$ performs the best, getting success in all the 10 examples. We further study the $X$ and $Y$ metrics, also provided in Table \ref{tab:examples}. Notably, $X_2$ is less strict in assessing TPs, assessing all DOs as true positives. In $\rm BEP_2$, $Y_2$ consequently plays the role of suitable metric, providing correct assessment for all the 10 examples. $Y_1$ is only slightly poorer than $Y_2$, providing 8 correct assessments out of 10. Thus, the role of bottom edge in correct assessment is verified.

\begin{table*}[t]
  \centering
  \caption{\small Qualitative comparison of metrics for examples in Fig. \ref{fig:distance}(b) is given here. The thresholds used for determining TPs are given in parentheses. For BEPs, ($x_0,y_0$) are given. The number of successes is the number of times a metric assesses the example as acceptable for maritime CV (i.e. number of matches with the maritime CV row).} \label{tab:examples} \vspace{-1mm}
  \scalebox{1}{
  \begin{tabular}{|l|c|c|c|c|c|c|c|c|c|c|c|}
     \hline
     Example & 1 & 2 & 3 & 4 & 5 & 6 & 7 & 8 & 9 & 10 & Number of Successes\\
     \hline
     Maritime CV           & \cellcolor{green}{TP} & \cellcolor{green}{TP}
                    & \cellcolor{green}{TP} & \cellcolor{red}{FP}
                    & \cellcolor{green}{TP} & \cellcolor{green}{TP}
                    & \cellcolor{red}{FP} & \cellcolor{red}{FP}
                    & \cellcolor{red}{FP} & \cellcolor{red}{FP}
                    & Not applicable\\
    \hline
    \hline
     IOU (0.5)      & \cellcolor{green}{TP} & \cellcolor{red}{FP}
                    & \cellcolor{red}{FP} & \cellcolor{green}{TP}
                    & \cellcolor{red}{FP} & \cellcolor{red}{FP}
                    & \cellcolor{red}{FP} & \cellcolor{green}{TP}
                    & \cellcolor{red}{FP} & \cellcolor{green}{TP}
                    & 3 \\
     Dice (0.5)     & \cellcolor{green}{TP} & \cellcolor{red}{FP}
                    & \cellcolor{red}{FP} & \cellcolor{green}{TP}
                    & \cellcolor{green}{TP} & \cellcolor{red}{FP}
                    & \cellcolor{green}{TP} & \cellcolor{green}{TP}
                    & \cellcolor{green}{TP} & \cellcolor{green}{TP}
                    & 2\\
     IOG (0.5)      & \cellcolor{green}{TP} & \cellcolor{red}{FP}
                    & \cellcolor{red}{FP} & \cellcolor{red}{FP}
                    & \cellcolor{green}{TP} & \cellcolor{red}{FP}
                    & \cellcolor{green}{TP} & \cellcolor{green}{TP}
                    & \cellcolor{red}{FP} & \cellcolor{green}{TP}
                    & 4\\
     $\rm BEP_1$ (0.7,0.75)
                    & \cellcolor{green}{TP} & \cellcolor{green}{TP}
                    & \cellcolor{green}{TP} & \cellcolor{red}{FP}
                    & \cellcolor{red}{FP} & \cellcolor{red}{FP}
                    & \cellcolor{red}{FP} & \cellcolor{green}{TP}
                    & \cellcolor{green}{TP} & \cellcolor{red}{FP}
                    & 6\\
     $\rm BEP_2$ (0.7,0.75)
                    & \cellcolor{green}{TP} & \cellcolor{green}{TP}
                    & \cellcolor{green}{TP} & \cellcolor{red}{FP}
                    & \cellcolor{green}{TP} & \cellcolor{green}{TP}
                    & \cellcolor{red}{FP} & \cellcolor{red}{FP}
                    & \cellcolor{red}{FP} & \cellcolor{red}{FP}
                    & 10\\
     \hline
     \hline
     $X_1$ (0.7)
                    & \cellcolor{green}{TP} & \cellcolor{green}{TP}
                    & \cellcolor{green}{TP} & \cellcolor{green}{TP}
                    & \cellcolor{red}{FP} & \cellcolor{red}{FP}
                    & \cellcolor{green}{TP} & \cellcolor{green}{TP}
                    & \cellcolor{green}{TP} & \cellcolor{green}{TP}
                    & 3\\
     $X_2$ (0.7)
                    & \cellcolor{green}{TP} & \cellcolor{green}{TP}
                    & \cellcolor{green}{TP} & \cellcolor{green}{TP}
                    & \cellcolor{green}{TP} & \cellcolor{green}{TP}
                    & \cellcolor{green}{TP} & \cellcolor{green}{TP}
                    & \cellcolor{green}{TP} & \cellcolor{green}{TP}
                    & 5\\
    \hline
    $Y_1$ (0.75)
                    & \cellcolor{green}{TP} & \cellcolor{green}{TP}
                    & \cellcolor{green}{TP} & \cellcolor{red}{FP}
                    & \cellcolor{green}{TP} & \cellcolor{green}{TP}
                    & \cellcolor{red}{FP} & \cellcolor{green}{TP}
                    & \cellcolor{green}{TP} & \cellcolor{red}{FP}
                    & 8\\
     $Y_2$ (0.75)
                    & \cellcolor{green}{TP} & \cellcolor{green}{TP}
                    & \cellcolor{green}{TP} & \cellcolor{red}{FP}
                    & \cellcolor{green}{TP} & \cellcolor{green}{TP}
                    & \cellcolor{red}{FP} & \cellcolor{red}{FP}
                    & \cellcolor{red}{FP} & \cellcolor{red}{FP}
                    & 10\\
    \hline
   \end{tabular}
   } \vspace{-2mm}
\end{table*}

\begin{table*}[t]
\caption{\small List of background subtraction methods is presented here. The methods are grouped according to the central concept behind them. The best results of each group appear in Table \ref{tab:recall}. The number of methods in each group is indicated in $\{\}$.}\label{tab:methods} \vspace{-1mm}
  \centering
  \scalebox{1}{
  \begin{tabular}{|p{40mm}|p{120mm}|}
    \hline
    \textbf{Group} & \textbf{Methods in the group}\\ \hline \hline
    Spatio-temporal filters (STF) - $\{4\}$ & Temporal mean (TM) \cite{lai1998fast}, Prati's median (PM) \cite{calderara2006reliable}, adaptive median (AM) \cite{mcfarlane1995segmentation}, $\sigma-\Delta$ BGS \cite{manzanera2007new}\\ \hline
    Gaussian models (GM) - $\{8\}$ & Simple Gaussian (SG) \cite{benezeth2008review}, Gaussian average (GA) \cite{wren1997pfinder}, Grimson's Gaussian mixture model (GMM) \cite{stauffer1999adaptive}, Zivkovic's adaptive GMM (AGMM) \cite{zivkovic2004improved}, mixture of Gaussians (MoG) \cite{sobral2014comprehensive}, fuzzy Gaussian (FG) \cite{sigari2008fuzzy}, type-2 fuzzy GMM - uncertain mean (T2FUM) \cite{zhao2012fuzzy}, type-2 fuzzy GMM - uncertain variance (T2FUV) \cite{zhao2012fuzzy} \\ \hline
    Kernel models (KM) - $\{2\}$ & Kernel density estimation (KDE) \cite{elgammal2000non}, VuMeter \cite{goya2006vehicle} \\ \hline
    Self organizing maps (SOM) - $\{2\}$ & Adaptive self organizing maps (ASOM) \cite{maddalena2010fuzzy}, fuzzy ASOM (FASOM) \cite{maddalena2010fuzzy} \\ \hline
     Low rank and sparsity (LRS) - $\{15\}$ & Eigen-background (EB) \cite{oliver2000bayesian}, active subspace (AS) robust principal component analysis (RPCA) \cite{liu2012active}, fast (F) principal component pursuit (PCP) \cite{rodriguez2013fast}, Reimanian robust (R2) PCP \cite{hintermuller2015robust}, MoG-RPCA \cite{zhao2014robust}, non-convex (NC) RPCA \cite{kang2015robust}, Grassman average \cite{hauberg2014grassmann}, greedy semi-soft go decomposition (GreGoDec) \cite{zhou2013greedy}, orthogonal rank-one matrix pursuit (OR1MP) \cite{wang2015orthogonal}, Grassmannian rank-one update subspace estimation (GROUSE) \cite{balzano2010online}, low-rank matrix completion by Riemannian optimization (LRGeomCG) \cite{vandereycken2013low}, non-negative matrix factorization (NMF) with sparse matrix (LS2) \cite{ji2013discriminative}, Deep semi NMF (DSNMF) \cite{trigeorgis2017deep}, alternating direction method of multipliers (ADMM) \cite{boyd2011distributed}, robust orthonormal subspace learning (ROSL) \cite{shu2014robust}\\ \hline
    Texture, color, and regions (TCR) - $\{5\}$ & Texture BGS (TBGS) \cite{heikkila2006texture}, independent multimodal background subtraction (IMBS) \cite{bloisi2012independent}, multicue \cite{noh2012new}, local binary similarity segmenter (LOBSTER) \cite{st2014improving}, self-balanced sensitivity segmenter (SuBSENSE) \cite{st2014flexible} \\
    \hline
  \end{tabular}
  }\vspace{-2mm}
\end{table*}

{The general criteria of assessing the pixel-based semantic segmentation are the same for maritime CV, where distance and span of an MV are important considerations. The bottom most pixels in semantic segmentations, which also form the bottom edge in a bounding box, are the most important determinant of the distance. The widest span of the semantic segmentation, which also forms the width of a bounding box, is the determinant of the span of the vessel. Therefore, the concept of BEPs is generally applicable to semantic segmentation as well.}

\section{Experiments and results}\label{sec:results}

Detection of MVs in a maritime environment falls under the ensemble problem set of `detection in dynamic background'. CV methods solve it by modeling and subtracting the dynamic background, followed by segmentation of the foreground \cite{bouwmans2014traditional,bouwmans2014background}. The dataset and the dynamic background subtraction methods used here are described below. We consider deep learning also for detection of MVs. These details are presented, followed by quantitative and qualitative results.

\paragraph{Dataset}
We use on-shore (fixed camera) visible range maritime videos from the maritime dataset, namely Singapore maritime dataset, published with \cite{prasad2017TITS}. There are 34 high-definition videos taken from Canon 70D cameras, Canon EF 70-300mm f/4-5.6 IS USM. The dataset has been captured at different times, such as before sunrise, at sunrise, at mid day, in the afternoon, in the evening, and 2 hours after sunset. We excluded the videos taken in haze and rain to avoid additional challenges. BBs of objects in each frame of the video are provided along with the dataset. Each BB is labeled with one of the following class labels: boat, buoy, ferry, flying bird/plane, kayak, sailboat, speed boat, vessel/ship, and others. {We have not included on-board videos for the reason we explain next. The motion of the vessel on which camera is mounted with respect to water and horizon presents additional challenges for dynamic background detection. The static background methods that use only current frame for background modeling are better candidates, but they have been shown to present extremely poor performance for maritime scenes \cite{prasad2017TITS}. 

\begin{figure*}[t]
  \centering
  \begin{tabular}{p{70mm} p{70mm}}
   \multicolumn{1}{c}{\small(a) Example frame 1} &
   \multicolumn{1}{c}{\small (b) Example frame 2} \\
   \multicolumn{1}{c}{\includegraphics[width=0.48\linewidth]{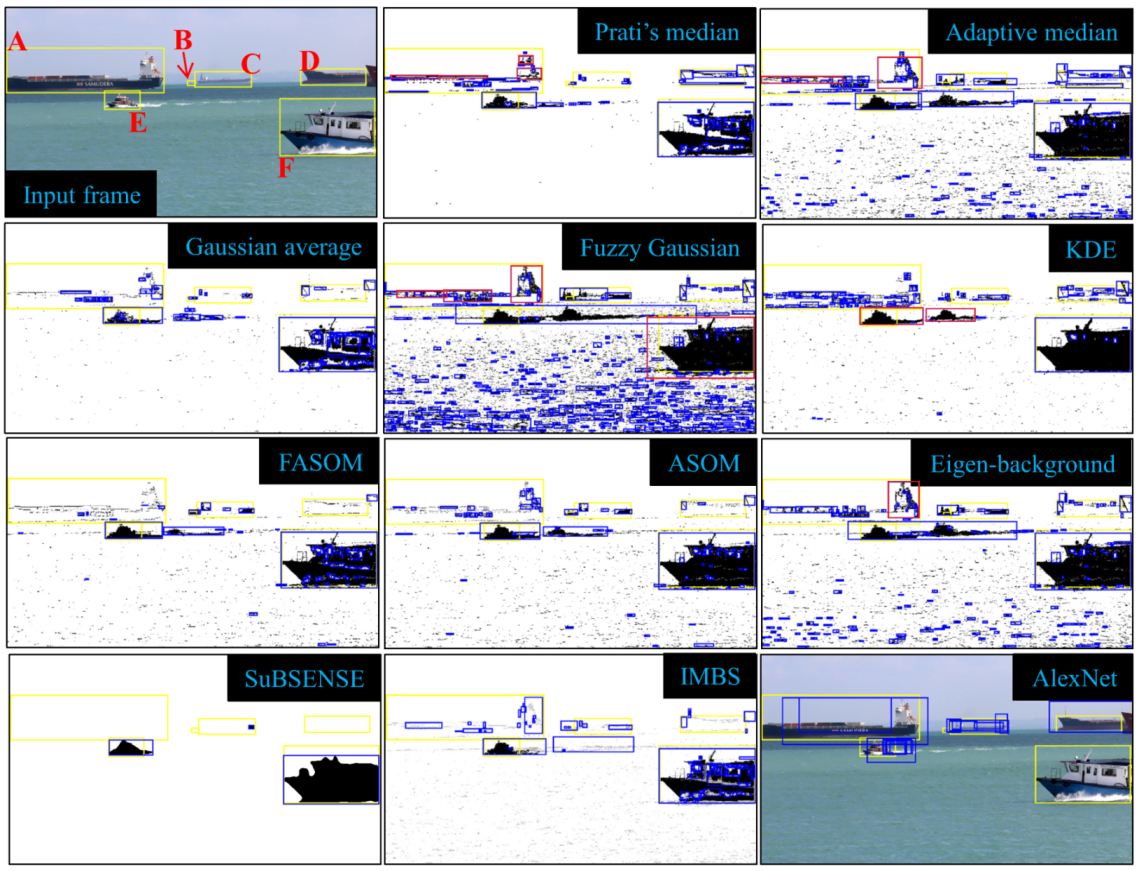}}&
   \multicolumn{1}{c}{\includegraphics[width=0.48\linewidth]{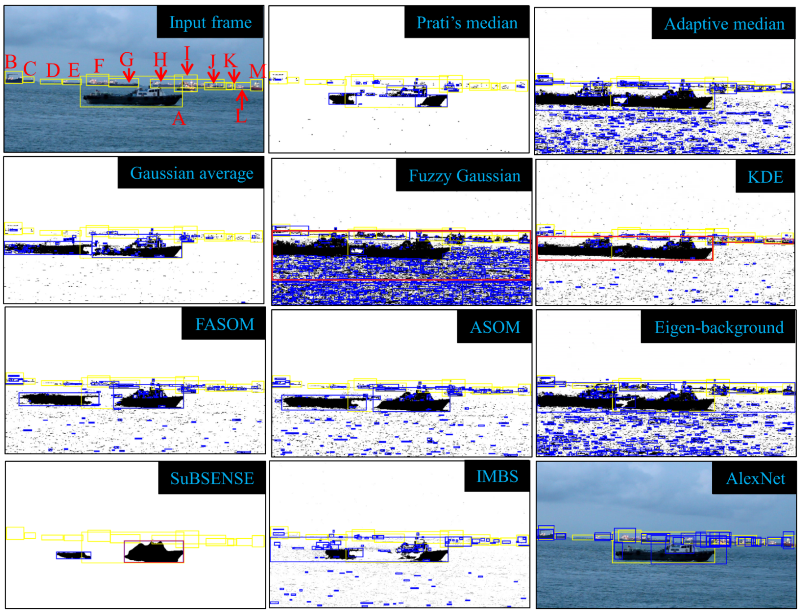}}\\
   \multicolumn{1}{c}{\small (c) Example frame 3}&
   \multicolumn{1}{c}{\small (d) Example frame 4}\\
   \multicolumn{1}{c}{\includegraphics[width=0.48\linewidth]{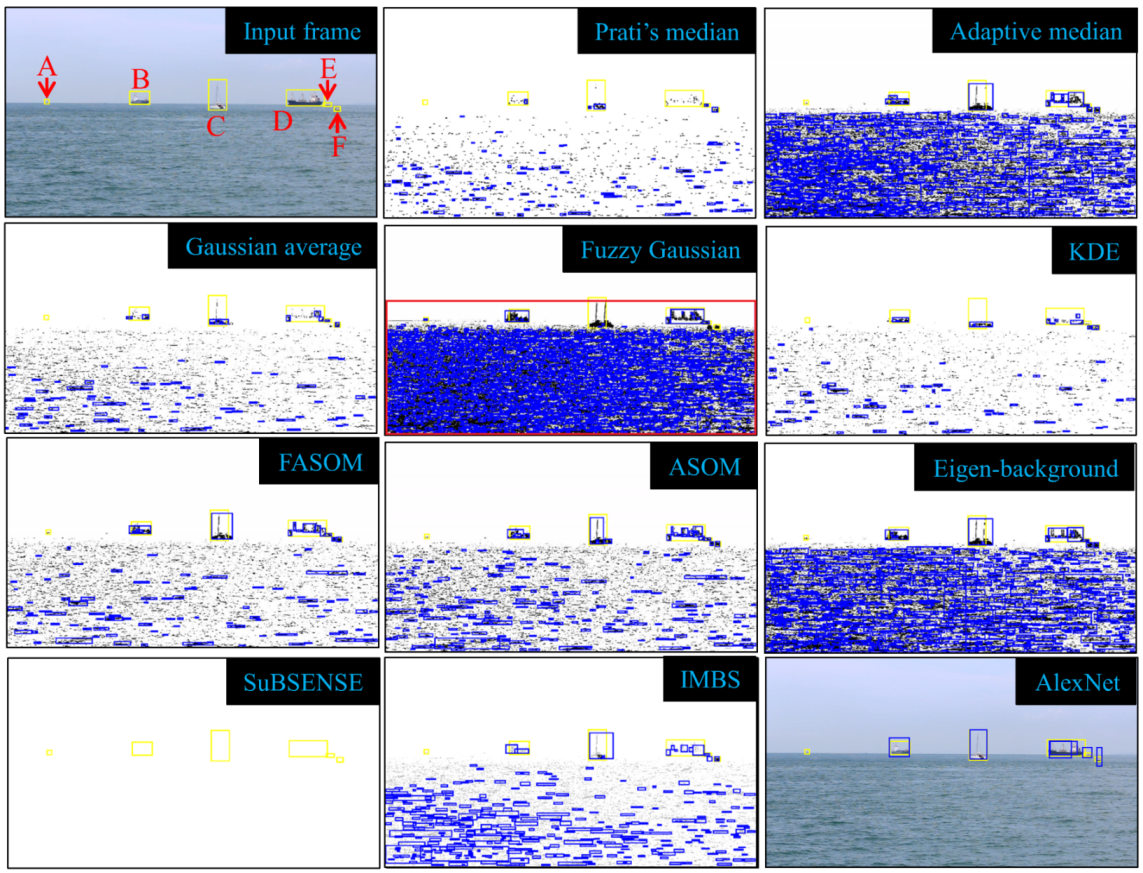}} &
   \multicolumn{1}{c}{\includegraphics[width=0.48\linewidth]{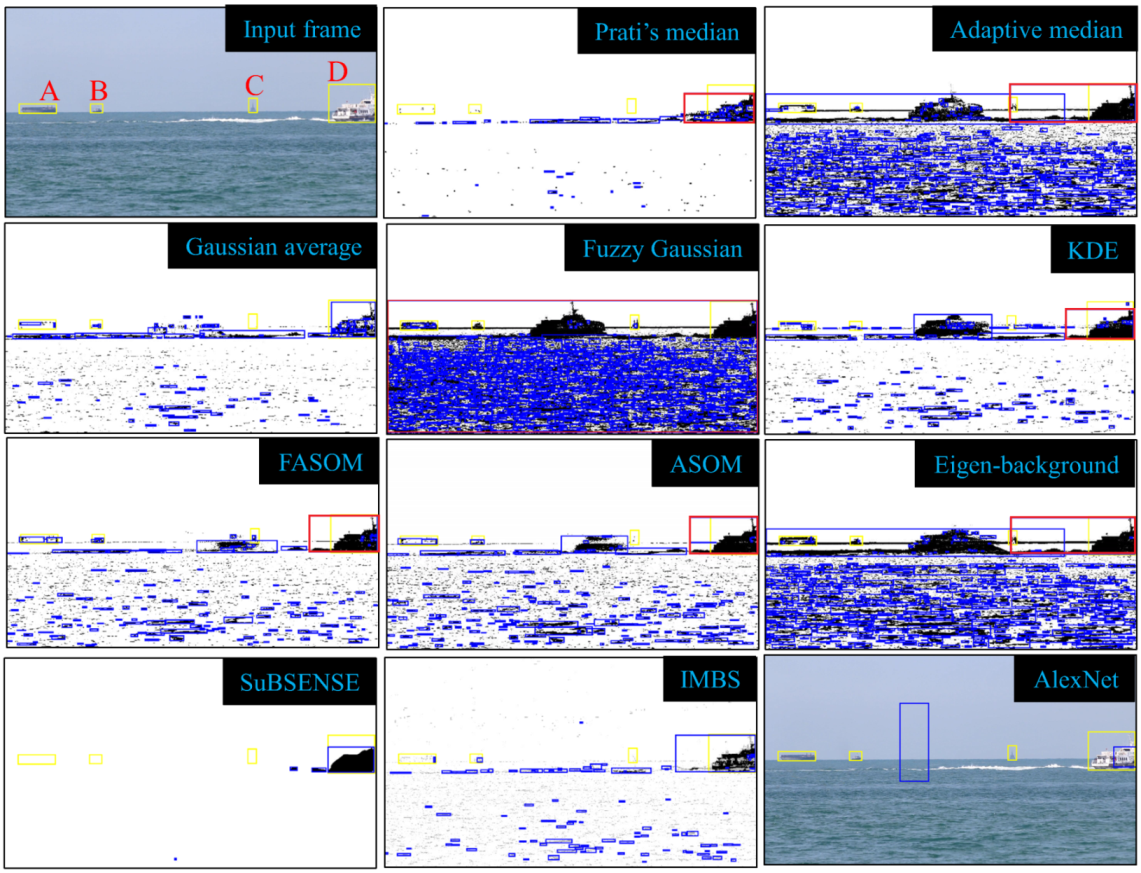}} \\

  \end{tabular}  \vspace{-1mm}
  \caption{\small Example results of CV methods for detection through dynamic BGS. The subtracted background appears white in the results of the methods. Ground truths: yellow BBs. Detected objects (foreground segmentations obtained after BGS): blue or red BBs. Red BBs: DOs referred in the text.} \label{fig:realexamples} \vspace{-4mm}
\end{figure*}

\paragraph{Dynamic background subtraction (BGS) methods tested}
We tested 22 BGS methods from the BGS library named bgslibrary \cite{sobral2014comprehensive,bgslibrary} and 14 BGS methods from the low rank and sparse (LRS) tools library name lrslibrary \cite{lrslibrary2015}. {The methods in BGS library are implemented in C++. The methods in LRS library are implemented in Matlab. All the methods were executed on Intel i7 6500 U @2.5 GHz desktop with 16 GB RAM and Linux platform.} Default parameters have been used for all the methods. Parameter tuning for achieving the best performance for each method is out of the scope of this work. {Yet, we note that fine tuning the control parameters for each method is likely to have a positive effect on the quality of detections, and is likely to impact all the metrics positively.} All detected BBs less than 20 pixels in any dimension are rejected as obviously spurious detections. We group the 36 methods into six broad categories based on their central concept. The groups and the methods in each of them are listed in Table \ref{tab:methods}. Among the 36 methods, only IMBS has been developed specifically for maritime scenes.

\paragraph{Regions with convolution neural network (R-CNN) features for detection using deep learning}
We conducted two experiments in deep learning. {These experiments were executed in Matlab on NVidia DGX-1 graphics server and Linux platform.} {The standard procedure of applying non-max suppression has not been used for the reason explained next. Many overlapping objects may be present in a maritime scene. Consider Fig. \ref{fig:realexamples}(b) for example. The GT bounding box of the vessel A overlaps with the GT bounding boxes of the vessels F, G, H, and I. Even if the DOs corresponding to them might be accurate, applying non-max suppression will result into lower recall because it will suppress either the DO of object A and other objects.} First, we randomly selected 20 videos from the dataset for training and trained R-CNN \cite{girshick2014rich} with AlexNet architecture. The results for this experiment were extremely poor and are not reported here. We attribute the poor performance to the challenging nature of the maritime scene and consider that maritime scenes may require camera and illumination specific training. In the second experiment, we formed the training dataset using every fifth frame of all the videos. The objective was to test if R-CNN can detect the objects it has been trained for. R-CNN trained on CIFAR-10 \cite{CIFAR10} performed poorly but R-CNN trained on ImageNet \cite{ImageNet} provided better results. {We note that R-CNN experiments may be considered to have unfair advantage over the other methods tested in this paper because the R-CNN experiments use training on a subset of images drawn from the main set itself.} We note also that use of R-CNN here \cite{girshick2014rich} is a first attempt of deep learning for maritime CV. Better suited approaches may be identified in the future. Some options include faster R-CNN \cite{ren2015faster}, long-term temporal convolution CNNs \cite{Varol2017}, networks on convolutional feature maps CNN \cite{Ren2017}.


\paragraph{Qualitative examples}
We consider four example frames, each taken from a different video of the dataset. The detection results of 10 BGS methods and R-CNN are shown in Fig. \ref{fig:realexamples}. The selected BGS methods are the ones that consistently outperform other methods in their groups either is precision or in recall. These methods are identified in Table \ref{tab:recall}. All BGS methods are ineffective in subtracting the background. In Fig. \ref{fig:realexamples}, all BGS methods except SuBSENSE detect false positive objects in the water background. This problem is more severe in frames 3 and 4, which show relatively more turbulent waters.

Consider fast moving objects in Fig. \ref{fig:realexamples}: E in frame 1, A in frame 2, and D in frame 4. Most methods generate phantom foreground for these objects, exceptions include Prati's median, SuBSENSE, and IMBS. Such phantoms may result into one wider detection or multiple individual detections, see KDE results for object A in frame 2 and object E in frame 1 for respective examples. These examples indicate a challenge not recognized in \cite{prasad2017TITS}. Dynamic BGS should incorporate large variations in the speeds of the vessels (both in the physical scales and the image scales) for avoiding phantom detections of fast vessels.

Wakes result in wider BBs in most methods for object D in frame 4. The detected spans of the fast moving objects and the objects with wakes are larger than the actual objects. For a fast moving object, information of minimum span and bottom edge is critically important for collision avoidance. It is acceptable, although not preferable, to interpret a larger span than the actual span. Thus, despite wider BBs, these detections are useful for collision avoidance. The BB of SuBSENSE corresponding to object A in frame 2 is comparatively less acceptable, since it underestimates the span of the vessel. IOU (0.5) estimates it as true positive, even though this detection indicates deficiency of SuBSENSE for collision avoidance. Also, note that fuzzy Gaussian BGS generates one significantly larger BB for each example frame, with the bottom edge of BB much below a GT's bottom edge. IOG detects it as a true positive, even though such detections are clearly deficient for collision avoidance.

Now, consider object A in frame 1 and objects B-D in frame 3. For these objects, several methods detect either the super-structure or the hull. Or, they break down the object into several smaller detections (note object A of frame 1). While the detected hulls indicate acceptable performance for collision avoidance, the detected super-structures or portions of the objects are unacceptable. BEPs are effective in assessing both these conditions appropriately.

Frame 2 presents an example of several occluded objects with small pixel foot prints. Different methods give varied results, several of them being useful for an initial estimate. This indicates potential for CV methods. However, suppression of false positive detections in water background is important for reasonable conclusion. At the same time, situations such as example 9 from Fig. \ref{fig:distance}(b) also occur in numerous places. See for example, the results of eigen-background and KDE for the example frame 2. Even with the BEP metrics, assessing them appropriately for collision avoidance in maritime CV is an open problem.

The results of R-CNN for the four example frames indicate that detections using R-CNN are better and less affected by wake. Moreover, DOs typically span both the hull and the super-structure. We note that the current implementation detects the same objects that it has been trained for, which is the reason for better quality of DOs. This approach is suitable only where environment specific training is feasible and practically useful.

\begin{table*}[t]
  \centering
  \caption{\small Precision and recall of CV methods for the maritime dataset. Best results for each group identified in Table \ref{tab:methods} are presented here. In each group, the methods that consistently give the best precision or recall for most assessment criteria are indicated in the bottom row.} \label{tab:recall} \vspace{-1mm}
  \begin{tabular}{|l|l|l|l|l|l|l|l|l|l|}
    \hline
    \multicolumn{10}{|c|}{Legend} \\ \hline
    \multicolumn{5}{|c|}{Precision} & \multicolumn{5}{|c|}{Recall} \\ \hline
    $\le 0.1$ & \cellcolor{red!10}$\le 0.2$ & \cellcolor{red!30}$\le 0.3$ & \cellcolor{red!50}$\le 0.4$ & \cellcolor{red!70}$\le 0.5$ & $\le 0.1$ & \cellcolor{blue!10}$\le 0.2$ & \cellcolor{blue!30}$\le 0.3$ & \cellcolor{blue!50}$\le 0.4$ & \cellcolor{blue!70}$\le 0.5$ \\
    \hline
  \end{tabular}
  \vspace{2mm}

 \scalebox{0.96}{
\begin{tabular}{|l|c|c||c|c|c|c|c|c|c||c|c|c|c|c|c|c|}
  \hline
  {} & \multicolumn{2}{c||}{parameters} & \multicolumn{7}{c||}{Precision} & \multicolumn{7}{c|}{Recall}\\ \hline
{} & $c_{0}$ or $x_{0}$ & $y_{0}$
& STF & GM & KM & SOM & LRS & TCR & {CNN}
& STF & GM & KM & SOM & LRS & TCR & {CNN}\\
\hline
\hline
\multirow{3}{*}{\parbox{10mm}{IOU $(c_0)$}} & 0.5 & {$-$}
&   0.01&0.00&0.01&0.01&0.00&\cellcolor{red!10}0.15 & \cellcolor{red!30}0.28
&   \cellcolor{blue!10}0.14&\cellcolor{blue!10}0.11&0.10&0.10&\cellcolor{blue!10}0.14&0.07 & \cellcolor{blue!50}0.41\\
 & 0.7 & {$-$}
& 0.00&0.00&0.00&0.00&0.00&0.05& \cellcolor{red!10}0.12
& 0.05&0.04&0.04&0.03&0.05&0.02 & \cellcolor{blue!10}0.18\\
 & 0.9 & {$-$}
& 0.00&0.00&0.00&0.00&0.00&0.00& 0.00
& 0.01&0.01&0.01&0.01&0.01&0.00 & 0.00\\
 \hline
\multirow{3}{*}{\parbox{10mm}{Dice $(c_0)$}} & 0.5 & {$-$}
&  0.01&0.01&0.01&0.01&0.00&\cellcolor{red!30}0.25 &\cellcolor{red!50}0.35
&  \cellcolor{blue!30}0.26&\cellcolor{blue!10}0.20&\cellcolor{blue!10}0.19&\cellcolor{blue!10}0.18&\cellcolor{blue!30}0.25&\cellcolor{blue!10}0.11 & \cellcolor{blue!50}0.51\\
 & 0.7 & {$-$}
 & 0.00&0.00&0.00&0.01&0.00&\cellcolor{red!10}0.14& \cellcolor{red!30}0.25
 & \cellcolor{blue!10}0.12&0.09&0.08&0.08&\cellcolor{blue!10}0.11&0.07 & \cellcolor{blue!50}0.37\\
 & 0.9 & {$-$}
 & 0.00&0.00&0.00&0.00&0.00&0.00& 0.03
 & 0.02&0.02&0.01&0.01&0.02&0.01 & 0.04\\
 \hline
\multirow{3}{*}{\parbox{10mm}{IOG $(c_0)$}} & 0.5 & {$-$}
& 0.01&0.01&0.01&0.01&0.07&\cellcolor{red!50}0.43 &\cellcolor{red!50}0.40
& \cellcolor{blue!50}0.32&\cellcolor{blue!30}0.30&\cellcolor{blue!10}0.20&\cellcolor{blue!10}0.19&\cellcolor{blue!50}0.32&\cellcolor{blue!10}0.19 &\cellcolor{blue!70}0.58\\
 & 0.7 & {$-$}
 & 0.01&0.01&0.01&0.01&0.07&\cellcolor{red!50}0.40& \cellcolor{red!50}0.32
 & \cellcolor{blue!30}0.24&\cellcolor{blue!30}0.26&\cellcolor{blue!10}0.14&\cellcolor{blue!10}0.13&\cellcolor{blue!30}0.25&\cellcolor{blue!10}0.17 & \cellcolor{blue!50}0.47\\
 & 0.9 & {$-$}
 & 0.00&0.00&0.00&0.00&0.07&\cellcolor{red!50}0.36& \cellcolor{red!10}0.17
 & \cellcolor{blue!10}0.15&\cellcolor{blue!10}0.19&0.09&0.07&\cellcolor{blue!10}0.17&\cellcolor{blue!10}0.16 & \cellcolor{blue!30}0.24\\
 \hline
\multirow{9}{*}{\parbox{10mm}{$\rm BEP_{1}$ $(x_0,y_0)$}}
 & $\sqrt{0.5}$ & 0.6
 & 0.01&0.01&0.01&0.01&0.00&\cellcolor{red!10}0.18 &\cellcolor{red!30}0.26
  & \cellcolor{blue!10}0.15&\cellcolor{blue!10}0.12&\cellcolor{blue!10}0.13&0.10&\cellcolor{blue!10}0.15&0.06 & \cellcolor{blue!50}0.38\\
 & $\sqrt{0.7}$ & 0.6
 & 0.00&0.01&0.01&0.01&0.00&\cellcolor{red!10}0.17 & \cellcolor{red!30}0.24
 & \cellcolor{blue!10}0.13&0.10&\cellcolor{blue!10}0.12&0.08&\cellcolor{blue!10}0.12&0.06 & \cellcolor{blue!50}0.35\\
 & $\sqrt{0.9}$ & 0.6
 & 0.00&0.00&0.00&0.00&0.00&\cellcolor{red!10}0.13 &\cellcolor{red!10}0.16
 & 0.08&0.07&0.08&0.06&0.08&0.04 & \cellcolor{blue!30}0.23\\ \cline{2-17}
 & $\sqrt{0.5}$ & 0.75
 & 0.00&0.00&0.00&0.00&0.00&\cellcolor{red!10}0.12 & \cellcolor{red!10}0.15
 & 0.09&0.07&0.08&0.05&0.09&0.04 & \cellcolor{blue!30}0.21\\
 & $\sqrt{0.7}$ & 0.75
 & 0.00&0.00&0.00&0.00&0.00&\cellcolor{red!10}0.11 & \cellcolor{red!10}0.14
 & 0.08&0.06&0.07&0.05&0.07&0.04 & \cellcolor{blue!10}0.20\\
 & $\sqrt{0.9}$ & 0.75
 & 0.00&0.00&0.00&0.00&0.00&0.10 & 0.09
 & 0.05&0.04&0.05&0.04&0.05&0.03 & \cellcolor{blue!10}0.13\\ \cline{2-17}
 & $\sqrt{0.5}$ & 0.9
 & 0.00&0.00&0.00&0.00&0.00&0.04 & 0.02
 & 0.03&0.03&0.03&0.02&0.03&0.01 & 0.03\\
 & $\sqrt{0.7}$ & 0.9
 & 0.00&0.00&0.00&0.00&0.00&0.04 & 0.02
 & 0.03&0.03&0.03&0.02&0.03&0.01 & 0.03\\
 & $\sqrt{0.9}$ & 0.9
 & 0.00&0.00&0.00&0.00&0.00&0.04 & 0.01
 & 0.02&0.02&0.03&0.02&0.02&0.01 & 0.02\\
 \hline \hline
\multirow{9}{*}{\parbox{10mm}{$\rm BEP_{2}$ $(x_0,y_0)$}}
 & $\sqrt{0.5}$ & 0.6
 & 0.01&0.01&0.02&0.01&0.00&\cellcolor{red!30}0.21 & \cellcolor{red!50} 0.33
 & \cellcolor{blue!50}0.38&\cellcolor{blue!50}0.31&\cellcolor{blue!30}0.27&\cellcolor{blue!30}0.23&\cellcolor{blue!50}0.38&\cellcolor{blue!10}0.12 & \cellcolor{blue!50} 0.49\\
 & $\sqrt{0.7}$ & 0.6
 & 0.01&0.01&0.01&0.01&0.00&\cellcolor{red!30}0.21 & \cellcolor{red!50}0.31
 & \cellcolor{blue!50}0.35&\cellcolor{blue!30}0.28&\cellcolor{blue!30}0.25&\cellcolor{blue!30}0.21&\cellcolor{blue!50}0.32&\cellcolor{blue!10}0.12 & \cellcolor{blue!50} 0.45\\
 & $\sqrt{0.9}$ & 0.6
 & 0.01&0.01&0.01&0.01&0.00&\cellcolor{red!10}0.17 & \cellcolor{red!30}0.21
 & \cellcolor{blue!30}0.25&\cellcolor{blue!10}0.20&\cellcolor{blue!10}0.20&\cellcolor{blue!10}0.16&\cellcolor{blue!30}0.25&0.09 & \cellcolor{blue!50} 0.31\\ \cline{2-17}
 & $\sqrt{0.5}$ & 0.75
 & 0.01&0.01&0.01&0.01&0.00&\cellcolor{red!10}0.16 & \cellcolor{red!30}0.26
 & \cellcolor{blue!50}0.33&\cellcolor{blue!30}0.26&\cellcolor{blue!30}0.23&\cellcolor{blue!10}0.19&\cellcolor{blue!50}0.33&0.10 & \cellcolor{blue!50}0.38\\
 & $\sqrt{0.7}$ & 0.75
 & 0.01&0.01&0.01&0.01&0.00&\cellcolor{red!10}0.16 & \cellcolor{red!30}0.23
 & \cellcolor{blue!30}0.30&\cellcolor{blue!30}0.24&\cellcolor{blue!30}0.21&\cellcolor{blue!10}0.17&\cellcolor{blue!30}0.30&0.10 & \cellcolor{blue!50}0.34\\
 & $\sqrt{0.9}$ & 0.75
 & 0.01&0.01&0.01&0.01&0.00&\cellcolor{red!10}0.13 & \cellcolor{red!10}0.15
 & \cellcolor{blue!30}0.23&\cellcolor{blue!10}0.18&\cellcolor{blue!10}0.18&\cellcolor{blue!10}0.14&\cellcolor{blue!30}0.23&0.08 & \cellcolor{blue!30}0.23\\ \cline{2-17}
 & $\sqrt{0.5}$ & 0.9
 & 0.01&0.01&0.01&0.01&0.00&0.09 & \cellcolor{red!10}0.16
 & \cellcolor{blue!30}0.26&\cellcolor{blue!30}0.21&\cellcolor{blue!10}0.18&\cellcolor{blue!10}0.14&\cellcolor{blue!30}0.27&0.08 & \cellcolor{blue!30}0.23\\
 & $\sqrt{0.7}$ & 0.9
 & 0.01&0.01&0.01&0.01&0.00&0.09 & \cellcolor{red!10}0.14
 & \cellcolor{blue!30}0.24&\cellcolor{blue!10}0.20&\cellcolor{blue!10}0.17&\cellcolor{blue!10}0.14&\cellcolor{blue!30}0.25&0.08 & \cellcolor{blue!10} 0.20\\
 & $\sqrt{0.9}$ & 0.9
 & 0.00&0.00&0.01&0.01&0.00&0.07 & 0.09
 & \cellcolor{blue!10}0.19&\cellcolor{blue!10}0.15&\cellcolor{blue!10}0.15&\cellcolor{blue!10}0.11&\cellcolor{blue!10}0.20&0.07 & \cellcolor{blue!10} 0.13\\
  \hline
  \hline
  \multirow{3}{*}{\parbox{10mm}{$\rm Y_{1}$ $(y_0)$}}
 & $-$ & 0.6
 & \cellcolor{red!10}0.12&\cellcolor{red!30}0.24&0.05&0.05&0.01&\cellcolor{red!70}0.59 & \cellcolor{red!70} 0.58
 & \cellcolor{blue!70}0.88&\cellcolor{blue!70}0.92&\cellcolor{blue!70}0.78&\cellcolor{blue!70}0.70&\cellcolor{blue!70}0.86&\cellcolor{blue!50}0.45 & \cellcolor{blue!70} 0.85\\
 & $-$ & 0.75
 & 0.09&\cellcolor{red!10}0.17&0.04&0.04&0.01&\cellcolor{red!70}0.53 & \cellcolor{red!70}0.55
 & \cellcolor{blue!70}0.81&\cellcolor{blue!70}0.87&\cellcolor{blue!70}0.72&\cellcolor{blue!70}0.63&\cellcolor{blue!70}0.80&\cellcolor{blue!50}0.37 & \cellcolor{blue!70}0.81\\
 & $-$ & 0.9
 & 0.05&0.07&0.03&0.03&0.01&\cellcolor{red!50}0.39 & \cellcolor{red!50}0.45
 & \cellcolor{blue!70}0.62&\cellcolor{blue!70}0.70&\cellcolor{blue!70}0.56&\cellcolor{blue!50}0.46&\cellcolor{blue!70}0.62&\cellcolor{blue!30}0.26 & \cellcolor{blue!70}0.65\\
   \hline
  \hline
  \multirow{3}{*}{\parbox{10mm}{$\rm Y_{2}$ $(y_0)$}}
   & $-$ & 0.6
 & 0.01&0.01&0.02&0.01&0.00&\cellcolor{red!30}0.23 & \cellcolor{red!50} 0.34
 & \cellcolor{blue!50}0.38& \cellcolor{blue!50}0.31&\cellcolor{blue!30}0.28&\cellcolor{blue!30}0.24&\cellcolor{blue!50}0.38&\cellcolor{blue!10}0.13 & \cellcolor{blue!50}0.49\\
 & $-$ & 0.75
 & 0.01&0.01&0.01&0.01&0.00&\cellcolor{red!10}0.17 & \cellcolor{red!30}0.24
 & \cellcolor{blue!30}0.30&\cellcolor{blue!30}0.24&\cellcolor{blue!30}0.22&\cellcolor{blue!10}0.18&\cellcolor{blue!30}0.30&0.10 & \cellcolor{blue!50}0.35\\
 & $-$ & 0.9
 & 0.00&0.01&0.01&0.01&0.00&0.08 & 0.09
& \cellcolor{blue!10}0.20&\cellcolor{blue!10}0.15&\cellcolor{blue!10}0.15&\cellcolor{blue!10}0.11&\cellcolor{blue!10}0.20&0.07 & \cellcolor{blue!10}0.13\\
 \hline
 \hline
\multicolumn{3}{|p{28mm}|}{{Consistently best}}
 & \rotatebox{90}{PM} & \rotatebox{90}{GA} & \rotatebox{90}{KDE} & \rotatebox{90}{FASOM} & \rotatebox{90}{EB} & \rotatebox{90}{SuBSENSE} & \rotatebox{90}{AlexNet}
 & \rotatebox{90}{AM} & \rotatebox{90}{FG} & \rotatebox{90}{KDE} & \rotatebox{90}{ASOM} &  \rotatebox{90}{EB} & \rotatebox{90}{IMBS} & \rotatebox{90}{AlexNet} \\
 \hline
\end{tabular}
}\vspace{-4mm}
\end{table*}

\paragraph{Quantitative results} We assess the true positive (TP) detections in all the frames of the all the videos in the dataset. The precision for the entire dataset is computed as the ratio of the total number of TPs to the total number of DOs. The recall is computed as the ratio of the total number of TPs to the total number of GTs. The assessment of TPs is performed using different assessment metrics and different threshold values for all of them. For IOU, Dice index, and IOG, we consider values 0.5, 0.7, and 0.9 for the threshold $c_0$. We note that IOU (0.5) is recommended in the well-known Pascal challenge \cite{everingham2015pascal}. The threshold $x_0$ for BEP$_1$ and BEP$_2$ is 1-dimensional analogue of $c_0$ for IOU and IOG, respectively. Thus, we use three values $\sqrt{0.5}$, $\sqrt{0.7}$, and $\sqrt{0.9}$ for $x_0$. We use three values 0.6, 0.75, and 0.9 for the threshold $y_0$. We include the results in which TPs are assessed using the $Y$ metrics alone. The precision and recall values of the six BGS groups identified in Table \ref{tab:methods} and the R-CNN are given in Table \ref{tab:recall}. The precision and recall values are color coded for easy visual interpretation.

TCR methods are more effective at background subtraction than the other methods (see results of SuBSENSE in Fig. \ref{fig:realexamples}). So, false positive detections due to water background are very few, leading to better precision than other methods. Also, precision values of SuBSENSE for BEP$_2$ metric are not poor considering that it was not developed specifically for the maritime domain. On the other hand, IMBS does not provide the best precision or recall even though it was developed specifically for the maritime domain. A reason could be that IMBS was developed for high mounted cameras in urban maritime, a setting different from the current dataset. The precision and recall results for R-CNN are expectedly better than the other approaches. However, noting that the R-CNN here detects the objects it has been trained for, the precision and recall should have been better. These clearly demonstrate the challenging nature of maritime CV.

The several false positives in most BGS methods (see Fig. \ref{fig:realexamples}) result in poor precision. Most methods have recall better than precision, with the exception of TCR methods. We also note that BEP$_2$ values are more encouraging than IOU, Dice Index, IOG, and BEP$_2$. The better suitability of BEP$_2$ was established in Table \ref{tab:examples}. Moreover, it is noted in Table \ref{tab:recall} that $Y_1$ is less selective about TPs. This puts the responsibility on $X_1$ for improving the selectivity of BEP$_1$. On the other hand, $Y_2$ is inherently more selective, as demonstrated by lower precision and recall values than $Y_1$. This directly helps in making BEP$_2$ selective.

We compare assessment metrics IOU(0.5) and BEP$_1(\sqrt{0.5},0.6)$, which correspond to most lenient threshold values. Recall values for BEP$_1(\sqrt{0.5},0.6)$ are better than IOU(0.5) in each group. For the most strict threshold values as well, recall values for BEP$_1(\sqrt{0.9},0.9)$ are better than IOU(0.9) in each group. The same can be inferred from the comparison of IOG and BEP$_2$, barring a few exceptions. Thus, although the conventional metrics indicate dismal performance of CV methods for maritime, the scene does not look so bleak when metrics designed for maritime domain are used. This highlights the need of both suitable metrics and dedicated CV solutions.

\section{Discussion}\label{sec:discussion}

We evaluated the existing metrics for assessing the quality of BB detections in the context of maritime CV. The unique needs of maritime CV imply that the current metrics are unsuitable. The proposed bottom edge proximity metrics, custom designed for maritime CV, provide a good starting point. However, there is a need to explore more options for assessing detections in maritime CV. Such assessment metrics would be strict in assessing the location of the bottom edges and minimum span of the BBs, suitable for assessing inaccurate detections due to occlusion, and tolerant for BB degradation in presence of wake or exclusion of super-structure in the detected BB. It is worth considering if the conventional BB labeling of GT is suitable for maritime CV. It should be explored if the GT of each vessel should comprise of GTs for hull, super-structure, and their union. An associated problem is to design assessment of detected BBs for such GT. Creating shape and pixel segmentations as ground truth for large videos needs to be explored. Detections and their assessment in the form of shape and pixel segmentations can be explored for new maritime CV methods.

Our preliminary study of 36 background subtraction methods and two R-CNN experiments shows a gap in CV techniques for maritime applications. Appropriate modeling of maritime background can reduce false positives and improve precision. Modeling wakes as background as well may allow stricter assessment of span (larger $x_0$) and thus better assessment of occlusions as well. Large range of speeds and sizes of maritime objects may require innovative approaches for learning background with adaptive time scales in local regions. Deep learning also holds significant promise. Our current experiments assume the luxury of environment specific training. A more generalizable deep learning framework for maritime is needed for practical maritime computer vision.

We note that the maritime computer vision is in a nascent stage at present. It is too early to decide on a suitable metric. A better convergence on these topics will emerge with further engagement of the CV community. The engagement can be through new diverse maritime datasets and maritime CV challenges similar to the PASCAL challenge \cite{everingham2015pascal} with goal towards autonomous maritime vehicle technology.



\bibliographystyle{ieeeTran}

\begin{thebibliography}{10}
\providecommand{\url}[1]{#1}
\csname url@samestyle\endcsname
\providecommand{\newblock}{\relax}
\providecommand{\bibinfo}[2]{#2}
\providecommand{\BIBentrySTDinterwordspacing}{\spaceskip=0pt\relax}
\providecommand{\BIBentryALTinterwordstretchfactor}{4}
\providecommand{\BIBentryALTinterwordspacing}{\spaceskip=\fontdimen2\font plus
\BIBentryALTinterwordstretchfactor\fontdimen3\font minus
  \fontdimen4\font\relax}
\providecommand{\BIBforeignlanguage}[2]{{%
\expandafter\ifx\csname l@#1\endcsname\relax
\typeout{** WARNING: IEEEtran.bst: No hyphenation pattern has been}%
\typeout{** loaded for the language `#1'. Using the pattern for}%
\typeout{** the default language instead.}%
\else
\language=\csname l@#1\endcsname
\fi
#2}}
\providecommand{\BIBdecl}{\relax}
\BIBdecl

\bibitem{Tu2017}
E.~Tu, G.~Zhang, L.~Rachmawati, E.~Rajabally, and G.-B. Huang, ``Exploiting
  {AIS} data for intelligent maritime navigation: A comprehensive survey from
  data to methodology,'' \emph{IEEE Transactions on Intelligent Transportation
  Systems}, vol.~PP, no.~99, pp. 1--24, 2017.

\bibitem{bloisi2009argos}
D.~Bloisi and L.~Iocchi, ``{ARGOS} {\-} {A} video surveillance system for boat
  traffic monitoring in {Venice},'' \emph{International Journal of Pattern
  Recognition and Artificial Intelligence}, vol.~23, no.~07, pp. 1477--1502,
  2009.


\bibitem{prasad2017TITS}
D.~K. Prasad, D.~Rajan, L.~Rachmawati, E.~Rajabaly, and C.~Quek, ``Video
  processing from electro-optical sensors for object detection and tracking in
  maritime environment: a survey,'' \emph{IEEE Transactions on Intelligent
  Transportation Systems}, vol.~18, no.~8, pp. 1993--2016, 2017.

\bibitem{withagen1999automatic}
P.~J. Withagen, K.~Schutte, A.~M. Vossepoel, and M.~G. Breuers, ``Automatic
  classification of ships from infrared ({FLIR}) images,'' in \emph{Signal
  Processing, Sensor Fusion, and Target Recognition VIII}, vol. 3720, 1999, pp.
  180--188.

\bibitem{Levandowsky1971}
M.~Levandowsky and D.~Winter, ``Distance between sets,'' \emph{Nature}, vol. 234, article no. 5323, 1971.

\bibitem{Altman1994}
D.~G.~Altman and J.~M.~Bland, ``Diagnostic tests. 1: Sensitivity and specificity,'' \emph{British Medical Journal}, vol. 308, no. 5323, pp. 1552, 1994.

\bibitem{Dice1945}
L.~R.~Dice, ``Measures of the amount of ecologic association between species,'' \emph{Ecology}, vol. 26, pp. 297--302, 1945.

\bibitem{cuzzocrea2017advanced}
A.~Cuzzocrea, E.~Mumolo, and G.~M. Grasso, ``Advanced pattern recognition from
  complex environments: a classification-based approach,'' \emph{Soft
  Computing}, pp. 1--16, 2017.

\bibitem{tversky1977features}
A.~Tversky, ``Features of similarity,'' \emph{Psychological Review}, vol.~84,
  no.~4, pp. 327--352, 1977.

\bibitem{prasad2018TITS}
D.~K. Prasad, C.~K. Prasath, D.~Rajan, L.~Rachmawati, E.~Rajabally, and
  C.~Quek, ``Object detection in maritime environment: Performance evaluation
  of background subtraction methods,'' \emph{IEEE Transactions on Intelligent
  Transportation Systems},vol. 22,  no.~5, pp. 1787--1802, 2019.

\bibitem{everingham2015pascal}
M.~Everingham, S.~A. Eslami, L.~Van~Gool, C.~K. Williams, J.~Winn, and
  A.~Zisserman, ``The pascal visual object classes challenge: A
  retrospective,'' \emph{International Journal of Computer Vision}, vol. 111,
  no.~1, pp. 98--136, 2015.

\bibitem{lai1998fast}
A.~H. Lai and N.~H. Yung, ``A fast and accurate scoreboard algorithm for
  estimating stationary backgrounds in an image sequence,'' in \emph{IEEE
  International Symposium on Circuits and Systems}, vol.~4, 1998, pp. 241--244.

\bibitem{calderara2006reliable}
S.~Calderara, R.~Melli, A.~Prati, and R.~Cucchiara, ``Reliable background
  suppression for complex scenes,'' in \emph{ACM International Workshop on
  Video Surveillance and Sensor Networks}, 2006, pp. 211--214.

\bibitem{mcfarlane1995segmentation}
N.~J. McFarlane and C.~P. Schofield, ``Segmentation and tracking of piglets in
  images,'' \emph{Machine Vision and Applications}, vol.~8, no.~3, pp.
  187--193, 1995.

\bibitem{manzanera2007new}
A.~Manzanera and J.~C. Richefeu, ``A new motion detection algorithm based on
  $\sigma$--$\delta$ background estimation,'' \emph{Pattern Recognition
  Letters}, vol.~28, no.~3, pp. 320--328, 2007.

\bibitem{benezeth2008review}
Y.~Benezeth, P.-M. Jodoin, B.~Emile, H.~Laurent, and C.~Rosenberger, ``Review
  and evaluation of commonly-implemented background subtraction algorithms,''
  in \emph{International Conference on Pattern Recognition}, 2008, pp. 1--4.

\bibitem{wren1997pfinder}
C.~R. Wren, A.~Azarbayejani, T.~Darrell, and A.~P. Pentland, ``Pfinder:
  Real-time tracking of the human body,'' \emph{IEEE Transactions on Pattern
  Analysis and Machine Intelligence}, vol.~19, no.~7, pp. 780--785, 1997.

\bibitem{stauffer1999adaptive}
C.~Stauffer and W.~E.~L. Grimson, ``Adaptive background mixture models for
  real-time tracking,'' in \emph{IEEE Conference on Computer Vision and Pattern
  Recognition}, vol.~2, 1999, pp. 246--252.

\bibitem{zivkovic2004improved}
Z.~Zivkovic, ``Improved adaptive gaussian mixture model for background
  subtraction,'' in \emph{International Conference on Pattern Recognition},
  vol.~2, 2004, pp. 28--31.

\bibitem{sobral2014comprehensive}
A.~Sobral and A.~Vacavant, ``A comprehensive review of background subtraction
  algorithms evaluated with synthetic and real videos,'' \emph{Computer Vision
  and Image Understanding}, vol. 122, pp. 4--21, 2014.

\bibitem{sigari2008fuzzy}
M.~H. Sigari, N.~Mozayani, and H.~Pourreza, ``Fuzzy running average and fuzzy
  background subtraction: concepts and application,'' \emph{International
  Journal of Computer Science and Network Security}, vol.~8, no.~2, pp.
  138--143, 2008.

\bibitem{zhao2012fuzzy}
Z.~Zhao, T.~Bouwmans, X.~Zhang, and Y.~Fang, ``A fuzzy background modeling
  approach for motion detection in dynamic backgrounds,'' in \emph{Multimedia
  and Signal Processing}, 2012, pp. 177--185.

\bibitem{elgammal2000non}
A.~Elgammal, D.~Harwood, and L.~Davis, ``Non-parametric model for background
  subtraction,'' in \emph{European Conference on Computer Vision}, 2000, pp.
  751--767.

\bibitem{goya2006vehicle}
Y.~Goya, T.~Chateau, L.~Malaterre, and L.~Trassoudaine, ``Vehicle trajectories
  evaluation by static video sensors,'' in \emph{IEEE Intelligent
  Transportation Systems Conference}, 2006, pp. 864--869.

\bibitem{maddalena2010fuzzy}
L.~Maddalena and A.~Petrosino, ``A fuzzy spatial coherence-based approach to
  background/foreground separation for moving object detection,'' \emph{Neural
  Computing and Applications}, vol.~19, no.~2, pp. 179--186, 2010.

\bibitem{oliver2000bayesian}
N.~M. Oliver, B.~Rosario, and A.~P. Pentland, ``A bayesian computer vision
  system for modeling human interactions,'' \emph{IEEE Transactions on Pattern
  Analysis and Machine Intelligence}, vol.~22, no.~8, pp. 831--843, 2000.

\bibitem{liu2012active}
G.~Liu and S.~Yan, ``Active subspace: Toward scalable low-rank learning,''
  \emph{Neural Computation}, vol.~24, no.~12, pp. 3371--3394, 2012.

\bibitem{rodriguez2013fast}
P.~Rodriguez and B.~Wohlberg, ``Fast principal component pursuit via
  alternating minimization,'' in \emph{IEEE International Conference on Image
  Processing}, 2013, pp. 69--73.

\bibitem{hintermuller2015robust}
M.~Hinterm{\"u}ller and T.~Wu, ``Robust principal component pursuit via inexact
  alternating minimization on matrix manifolds,'' \emph{Journal of Mathematical
  Imaging and Vision}, vol.~51, no.~3, pp. 361--377, 2015.

\bibitem{zhao2014robust}
Q.~Zhao, D.~Meng, Z.~Xu, W.~Zuo, and L.~Zhang, ``Robust principal component
  analysis with complex noise,'' in \emph{International Conference on Machine
  Learning}, 2014, pp. 55--63.

\bibitem{kang2015robust}
Z.~Kang, C.~Peng, and Q.~Cheng, ``Robust pca via nonconvex rank
  approximation,'' in \emph{IEEE International Conference on Data Mining},
  2015, pp. 211--220.

\bibitem{hauberg2014grassmann}
S.~Hauberg, A.~Feragen, and M.~J. Black, ``Grassmann averages for scalable
  robust pca,'' in \emph{IEEE Conference on Computer Vision
  and Pattern Recognition}, 2014, pp. 3810--3817.

\bibitem{zhou2013greedy}
T.~Zhou and D.~Tao, ``Greedy bilateral sketch, completion \& smoothing,'' in
  \emph{International Conference on Artificial Intelligence and Statistics},
  2013.

\bibitem{wang2015orthogonal}
Z.~Wang, M.-J. Lai, Z.~Lu, W.~Fan, H.~Davulcu, and J.~Ye, ``Orthogonal rank-one
  matrix pursuit for low rank matrix completion,'' \emph{SIAM Journal on
  Scientific Computing}, vol.~37, no.~1, pp. A488--A514, 2015.

\bibitem{balzano2010online}
L.~Balzano, R.~Nowak, and B.~Recht, ``Online identification and tracking of
  subspaces from highly incomplete information,'' in \emph{Annual Allerton
  Conference on Communication, Control, and Computing}, 2010, pp. 704--711.

\bibitem{vandereycken2013low}
B.~Vandereycken, ``Low-rank matrix completion by riemannian optimization,''
  \emph{SIAM Journal on Optimization}, vol.~23, no.~2, pp. 1214--1236, 2013.

\bibitem{ji2013discriminative}
Y.~Ji and J.~Eisenstein, ``Discriminative improvements to distributional
  sentence similarity,'' in \emph{Conference on Empirical Methods in Natural
  Language Processing}, 2013, pp. 891--896.

\bibitem{trigeorgis2017deep}
G.~Trigeorgis, K.~Bousmalis, S.~Zafeiriou, and B.~W. Schuller, ``A deep matrix
  factorization method for learning attribute representations,'' \emph{IEEE
  Transactions on Pattern Analysis and Machine Intelligence}, vol.~39, no.~3,
  pp. 417--429, 2017.

\bibitem{boyd2011distributed}
S.~Boyd, N.~Parikh, E.~Chu, B.~Peleato, J.~Eckstein \emph{et~al.},
  ``Distributed optimization and statistical learning via the alternating
  direction method of multipliers,'' \emph{Foundations and Trends in Machine
  learning}, vol.~3, no.~1, pp. 1--122, 2011.

\bibitem{shu2014robust}
X.~Shu, F.~Porikli, and N.~Ahuja, ``Robust orthonormal subspace learning:
  Efficient recovery of corrupted low-rank matrices,'' in \emph{IEEE Conference
  on Computer Vision and Pattern Recognition}, 2014, pp. 3874--3881.

\bibitem{heikkila2006texture}
M.~Heikkila and M.~Pietikainen, ``A texture-based method for modeling the
  background and detecting moving objects,'' \emph{IEEE Transactions on Pattern
  Analysis and Machine Intelligence}, vol.~28, no.~4, pp. 657--662, 2006.

\bibitem{bloisi2012independent}
D.~Bloisi and L.~Iocchi, ``Independent multimodal background subtraction,'' in
  \emph{International Conference on Computational Modeling of Objects Presented
  in Images: Fundamentals, Methods and Applications}, 2012, pp. 39--44.

\bibitem{noh2012new}
S.~Noh and M.~Jeon, ``A new framework for background subtraction using multiple
  cues,'' in \emph{Asian Conference on Computer Vision}, 2012, pp. 493--506.

\bibitem{st2014improving}
P.-L. St-Charles and G.-A. Bilodeau, ``Improving background subtraction using
  local binary similarity patterns,'' in \emph{IEEE Winter Conference on
  Applications of Computer Vision}, 2014, pp. 509--515.

\bibitem{st2014flexible}
P.-L. St-Charles, G.-A. Bilodeau, and R.~Bergevin, ``Flexible background
  subtraction with self-balanced local sensitivity,'' in \emph{IEEE Conference
  on Computer Vision and Pattern Recognition Workshops}, 2014, pp. 408--413.

\bibitem{bgslibrary}
\BIBentryALTinterwordspacing
A.~Sobral, ``{BGSLibrary}: An opencv c++ background subtraction library,''
  2013, pp. 1--16. [Online]. Available:
  \url{https://github.com/andrewssobral/bgslibrary}
\BIBentrySTDinterwordspacing

\bibitem{lrslibrary2015}
A.~Sobral, T.~Bouwmans, and E.-h. Zahzah, ``{LRSLibrary}: Low-rank and sparse
  tools for background modeling and subtraction in videos,'' in \emph{Robust
  Low-Rank and Sparse Matrix Decomposition: Applications in Image and Video
  Processing}.\hskip 1em plus 0.5em minus 0.4em\relax CRC Press, Taylor and
  Francis Group.

\bibitem{girshick2014rich}
R.~Girshick, J.~Donahue, T.~Darrell, and J.~Malik, ``Rich feature hierarchies
  for accurate object detection and semantic segmentation,'' in \emph{IEEE
  conference on Computer Vision and Pattern Recognition}, 2014, pp. 580--587.

\bibitem{ren2015faster}
S.~Ren, K.~He, R.~Girshick, and J.~Sun, ``Faster {R-CNN}: Towards real-time
  object detection with region proposal networks,'' in \emph{Advances in Neural
  Information Processing Systems}, 2015, pp. 91--99.

\bibitem{Varol2017}
G.~Varol, I.~Laptev, and C.~Schmid, ``Long-term temporal convolutions for
  action recognition,'' \emph{IEEE Transactions on Pattern Analysis and Machine
  Intelligence}, vol.~PP, no.~99, pp. 1--1, 2017.

\bibitem{Ren2017}
S.~Ren, K.~He, R.~Girshick, X.~Zhang, and J.~Sun, ``Object detection networks
  on convolutional feature maps,'' \emph{IEEE Transactions on Pattern Analysis
  and Machine Intelligence}, vol.~39, no.~7, pp. 1476--1481, 2017.

\bibitem{bouwmans2014traditional}
T.~Bouwmans, ``Traditional and recent approaches in background modeling for foreground detection: An overview,'' \emph{Computer Science Review}, vol. 11, pp. 31-66,2014.

\bibitem{bouwmans2014background}
T.~Bouwmans, F.~Porikli, B.~H{\"o}ferlin and A.~Vacavant \emph{Background modeling and foreground detection for video surveillance}, CRC Press, 2014.

\bibitem{CIFAR10}
A.~Krizhevsky and G.~Hinton ``Learning multiple layers of features from tiny images,'', vol. 1, no. 4. \emph{Technical report}, University of Toronto, 2009.

\bibitem{ImageNet}
J.~Deng, W.~Dong, et. al, ``Imagenet: A large-scale hierarchical image database,'' in \emph{IEEE
  conference on Computer Vision and Pattern Recognition}, 2009, pp. 248--255.


\end{thebibliography}

\vspace{-14mm}
\begin{IEEEbiography}[{\includegraphics[width=0.7in,height=1in,clip,keepaspectratio]{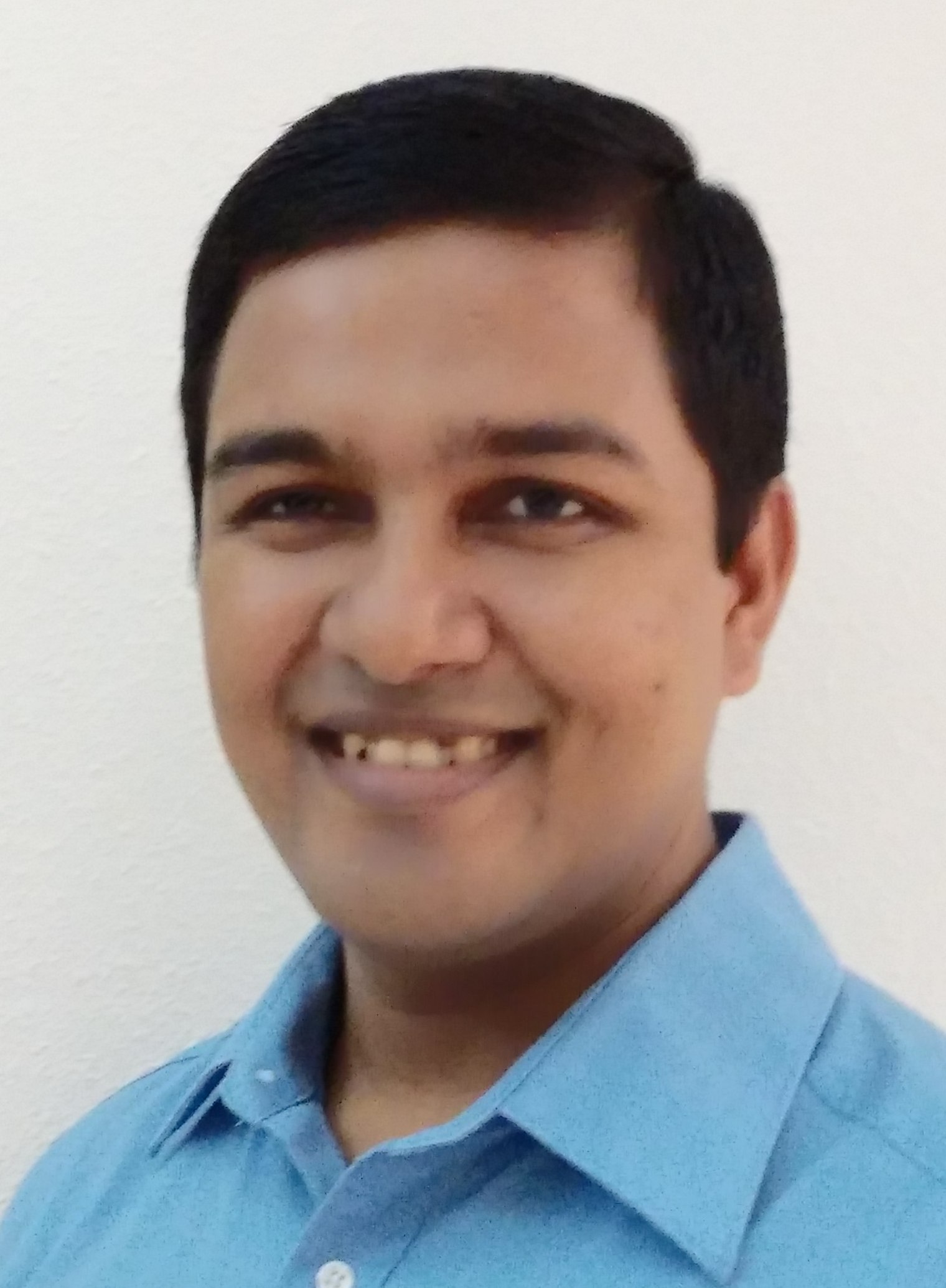}}]{Dilip K. Prasad} received the B.Tech. and Ph.D. degrees in computer science and engineering from the Indian Institute of Technology (ISM), Dhanbad, India, and Nanyang Technological University, Singapore, in 2003 and 2013, respectively. He is currently an associate professor at UiT The Arctic University of Norway. His current research interests include image processing, machine learning, and computer vision.
 \end{IEEEbiography}

\vspace{-15mm}
\begin{IEEEbiography}[{\includegraphics[width=0.7in,height=1in,clip,keepaspectratio]{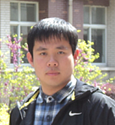}}] {Huixu Dong}  received the B.Sc degree in mechatronics engineering from Harbin Institute of Technology in China, in 2013 and  Ph.D. degree at Robotics Research Centre of Nanyang Technological University, Singapore, in 2018. Currently, he is a post-doctoral fellow in Robotics Institute of Carnegie Mellon University. His current research interests include robotic perception and grasp in unstructured environments, computer vision and robot-oriented artificial intelligence, the navigation of mobile robot and optimal design of robotic gripper.
 \end{IEEEbiography}

\vspace{-15mm}
\begin{IEEEbiography}[{\includegraphics[width=0.7in,height=1in,clip,keepaspectratio]{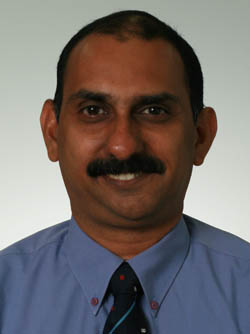}}] {Deepu Rajan} received the Bachelor of Engineering degree in electronics and communication engineering from the Birla Institute of Technology, Ranchi, India, the M.S. degree in electrical engineering from Clemson University, Clemson, SC, USA, and the Ph.D. degree from the Indian Institute of Technology, Mumbai, India. He is an Associate Professor with the School of Computer Engineering, Nanyang Technological University, Singapore. His current research interests include image processing, computer vision, and multimedia signal processing.
 \end{IEEEbiography}

\vspace{-14mm}
\begin{IEEEbiography} [{\includegraphics[width=0.7in,height=1in,clip,keepaspectratio]{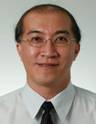}}] {Chai Quek} is with the School of Computer Engineering, Nanyang Technological University, Singapore. He received B.Sc. and Ph.D. degrees from Heriot-Watt University, Edinburgh, U.K. His research interests include neurocognitive informatics, biomedical engineering and computational finance. He has published over 250 international conference and journal papers. He has been invited as a Program Committee Member and reviewer for several conferences and journals, including IEEE TNN, TEvC, etc.
\end{IEEEbiography}
\end{document}